\documentclass[10pt,twocolumn,letterpaper]{article}

\usepackage{cvpr}
\usepackage{times}
\usepackage{epsfig}
\usepackage{graphicx}
\usepackage{amsmath}
\usepackage{amssymb}
\usepackage{textcomp}
\usepackage{url}

\usepackage{graphicx}
\usepackage{epstopdf}
\usepackage{epsfig}
\usepackage{bbm}
\usepackage{subfigure}
\usepackage[colorlinks,linkcolor=red,breaklinks=true]{hyperref}

\def\e{{\bf e}}

\def\0{{\bf 0}}
\def\1{{\bf 1}}

\def\etal{{\em et al.\/}\,}

\cvprfinalcopy 

\def\cvprPaperID{1005} 

\ifcvprfinal\fi
\begin{document}

\title{On The Stability of Video Detection and Tracking}

\author{ Hong Zhang\\
The Chinese University of Hong Kong\\
{\tt\small fykalviny@gmail.com}
\and
 Naiyan Wang\\
 TuSimple\\
{\tt\small winsty@gmail.com}
}

\maketitle

\begin{abstract}
In this paper, we study an important yet less explored aspect in video detection and tracking -- stability. Surprisingly, there is no prior work that tried to study it. As a result, we start our work by proposing a novel evaluation metric for video detection which considers both stability and accuracy. For accuracy, we extend the existing accuracy metric mean Average Precision (mAP). For stability, we decompose it into three terms: fragment error, center position error, scale and ratio error. Each error represents one aspect of stability. Furthermore, we demonstrate that the stability metric has low correlation with accuracy metric. Thus, it indeed captures a different perspective of quality. Lastly, based on this metric, we evaluate several existing methods for video detection and show how they affect accuracy and stability. We believe our work can provide guidance and solid baselines for future researches in the related areas.
\end{abstract}

\section{Introduction}

Object detection refers to the problem that localizes and classifies the objects of interest in an image. It serves as a fundamental task for many other high level tasks like human computer interaction. In its early stage, most researches focused on certain object detection, such as face~\cite{viola2004robust,rowley1998neural,osuna1997training}, hand~\cite{kolsch2004robust,eng2004a} or pedestrian~\cite{leibe2005pedestrian,oren1997pedestrian}, \etc. For general object detection, early methods such as Deformable Part Model (DPM)~\cite{dpm} relies on hand-crafted features and deliberately designed classifiers. Nowadays, with the advancement of deep learning technique, current state-of-the-art paradigm shifts to data driven end-to-end learning. Some representative methods include region based methods~\cite{rcnn,fastrcnn,fasterrcnn,spp} and direct regression methods~\cite{yolo,ssd}.

\begin{figure}[bpt]
\centering
\begin{tabular}{@{\hspace{0mm}}c}
\includegraphics[width=0.9\linewidth]{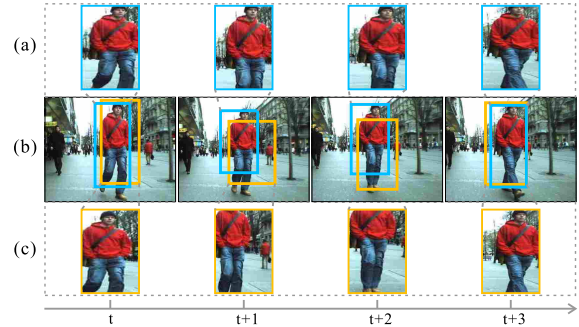}
\\
\end{tabular}
\caption{A pivot illustration to demonstrate the insufficiency of current VID evaluation. Here are the results of two detectors in one video with the \emph{same mAP}. (a) Stable trajectory. Though the detector misses the feet of the pedestrian, it is consistent on the region it includes, and almost centered. (b) Consecutive frames with results of two detectors. (c) Unstable trajectory. Predicted bounding boxes contain different parts of object and jitter.}
\label{fig:motivation}
\end{figure}

Although most works focus on still image detection, another equally important yet emerging setting is Video Detection (VID). Video detection could leverage the temporal context within consecutive frames to alleviate the intrinsic difficulties in single image detection, such as motion blurs and occlusions. To push this field forward, Imagenet LSVRC~\cite{imagenet} introduced the video detection challenge recently. In just two years, we have witnessed that the performance improved from 67.82 to 80.83 in mean Average Precision (mAP) rapidly. Despite its rapid development, we have observed that there are some disturbing facts regarding to the evaluation of VID algorithms. For example, in Fig.~\ref{fig:motivation}, two methods get the same results on this trajectory based on mAP. Nevertheless, (a) is obviously better than (c) for human judgment because (a) is stable around the person, however (c) jitters a lot. We would like to ask: \emph{What is the missing component in current VID evaluation?}



Our answer is stability. Currently, the evaluation of VID mainly based on the Intersection over Union (IoU) which aggregates the results of still images or tracklets. 
However, stability is always an important factor to consider in practice. For example, in an Advanced Driving Assistance System (ADAS), we need to utilize the change rate of bounding boxes to estimate the Time To Collision (TTC)~\cite{ttc} and relative speed~\cite{acc}, which is at the core of the safety warning system. If the center or scale of the bounding box jitters around the object of interest, it is obvious that the estimated TTC, speed and distance are inaccurate and unstable. Unfortunately, there is no evaluation metric to quantify such phenomenon properly.

In this paper, we highlight the importance of stability in video detection in addition to its accuracy counterpart. 
In particular, we move the paradigm of evaluation from bounding box centric to trajectory centric. It assesses the stability along each trajectory, which includes temporal continuity, center position stability, scale and ratio stability, respectively. Guided by the new evaluation metric, we also benchmark several existing methods to improve VID performance. The empirical results also reveal an interesting finding: The existing accuracy metric and proposed stability metric are less correlated. They actually capture different aspects of VID quality. We wish these methods could be served as effective baselines for future researches.

To summarize, our contributions are in the following three folds:
\begin{itemize}
\item We propose a novel evaluation metric to assess the performance of VID methods. The proposed metric considers a crucial yet usually ignored aspect of VID -- stability.
\item We empirically demonstrate that the stability metric has low correlation with existing accuracy metric, thus it is meaningful to evaluate both of them.
\item We evaluate some existing baselines under our new evaluation metric, and show trade-offs between them. We wish these benchmarks could lighten further research directions in the field.
\end{itemize}

\section{Related Work}
Object detection has a long history in computer vision community. Before deep learning age, the conventional methods heavily relied on hand-crafted features and carefully designed pipelines. One representative work is the Deformable Part Model (DPM)~\cite{dpm}. Afterwards, instead of using sliding window, researchers proposed to first generate object proposals that may contain objects, and then classify them into different categories. Widely used methods to generate proposals include those based on super-pixels grouping, \eg Selective Search~\cite{selectivesearch}, MCG~\cite{mcg} and those based on sliding window and edge features, \eg EdgeBoxes~\cite{edgeboxes}. 

Some early works tried to apply CNN to object detection include~\cite{overfeat,erhan2014scalable,szegedy2013deep}. However, their performance does not significantly outperform the conventional methods. Region CNN (RCNN)~\cite{rcnn} is a milestone in object detection. Briefly speaking, it adopts CNNs to extract features of region proposals and score them via learned classifiers. Subsequent works Fast RCNN~\cite{fastrcnn} and Faster RCNN~\cite{fasterrcnn} accelerate it by reusing existing feature maps.
To further speed up, several real-time methods~\cite{yolo,ssd} were also proposed. These methods cast detection into direct regression problem. Namely, the network jointly predicts bounding boxes and confidence scores in an end-to-end manner.

Beyond object detection in still images, video detection aims to detect objects in video sequences. Until recently, a few researchers have done several pioneering works. The keys of these methods lie in utilizing temporal context information. In~\cite{kai16tcnn,kang16object}, Kang \etal first tried to use object class correlation and motion propagation to reduce false positives and false negatives, and then trained a temporal convolution network to rescore detections based on the tubelets generated by visual tracking. Han \etal~\cite{seqnms} used the sequence Non-Maximal Suppression (NMS) to rescore the detection results. Very recently, Tripathi \etal~\cite{tripathi2016context} trained a Recurrent Neural Network (RNN) on the initial detections results to refine them. Unfortunately, all these methods are limited in post-processing stage. Few works tried to integrate the temporal context in an end-to-end manner. Interestingly, for a closely related area -- video segmentation, \cite{fayyaz2016stfcn, valipour2016recurrent} adopted the Convolutional Long Short Term Memory (Conv-LSTM) to capture both temporal and spatial information in one unified model. For the evaluation of video segmentation, a recent work~\cite{kundu2016feature} evaluated the temporal consistency as well as the accuracy in a single frame. We believe video detection is still at its early stage compared with video segmentation. There are still a lot of issues that need to be addressed in the future. Among those, stability is the foremost one that needs to be investigated.


Video detection is also a prerequisite for most MOT algorithms. 
MOT usually assumes the detections exist, and focus on associating and refining the detections across different frame. The MOT algorithms can be roughly categorized into two types: (Nearly) online algorithms~\cite{michael09robust, wongun13a, zia05mcmc} try to associate existing targets with detections in recent frames and output results immediately after receiving the input image. Meanwhile, offline algorithms~\cite{berclaz2011multiple, anton14continuous} read in all frames, and then output the results afterwards. They could utilize the later frames to refine the results in early frames, thus usually get better results. Note that our proposed metric can also be applied to MOT. It captures a different aspect of trajectory quality in addition to the popular CLEAR MOT metric~\cite{mota}. The readers could refer to Sec.~\ref{sec:mota} for more discussions.

\begin{figure}[bpt]
\centering
\begin{tabular}{@{\hspace{0mm}}c}
\includegraphics[width=0.85\linewidth]{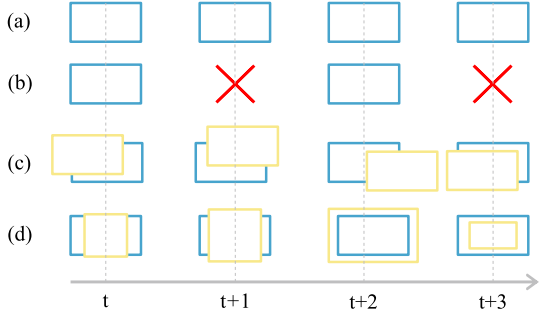}
\\
\end{tabular}
\caption{An illustration of four trajectories. (a) Ground-truth trajectory. (b) Interrupted trajectory. (c) Center position jittering trajectory. (d) Scale and ratio jittering trajectory.}
\label{fig:example}
\end{figure}
\section{Our Evaluation Metric}
In this section, we present our novel evaluation metric for VID. The metric consists of two parts: The first part is detection accuracy which evaluates whether the objects are precisely localized and classified. The other part considers the detection stability for each trajectory both temporally and spatially. Temporal stability is to measure the integrity of a trajectory, while spatial stability is to measure how much the detection bounding boxes jitter around the ground-truths in a trajectory. An illustration for the stability metric is presented in Fig.~\ref{fig:example}. We will elaborate the details in the following sections.

\subsection{Detection Accuracy}
For accuracy, we simply extend the existing metric in still image detection.
In each frame, the performance of a detector can be measured by the overlap rate between the ground-truths and the predicted bounding boxes. The overlap rate is defined as the area of the intersection of two bounding boxes over their union (IoU). Given an IoU threshold to be true positive, we can generate the precision and recall curve by varying the threshold of detection score. Average Precision (AP) is the Area Under Curve (AUC) of precision and recall curve. Then we vary the IoU threshold from 0 gradually to 1, and we calculate the AUC of AP curve as detection accuracy. A larger AUC value indicates better accuracy. If the dataset contains more than one class, we simply use the mean AUC (mAUC) over all classes as the final detection accuracy.

\subsection{Detection Stability}
To evaluate the stability, we need to assign each detection to a trajectory. If the algorithms output such association as in MOT, we can directly use them. If not, we first apply the Hungarian algorithm~\cite{hungarian} to find the best possible matching between the output detections and ground-truths. IoUs between them are treated as the weights of the bipartite graph. The detections that are not paired to any ground-truth are excluded in stability evaluation. Eq. (\ref{eq:temporal}) is the formulation for detection stability:
\begin{equation}
\Phi = E_F + E_C + E_R,
\label{eq:temporal}
\end{equation}
where $E_F$ is the fragment error, $E_C$ is the center position error, $E_R$ is the scale and ratio error. Same as in detection accuracy, we also change the IoU threshold of true positive to examine the trends between the errors and thresholds. For each IoU threshold, we can similarly draw its corresponding error \vs recall curve by varying the threshold of detection score. We use the AUC of this curve as the stability error at a certain IoU threshold. At last, we treat the AUC of the IoU threshold \vs stability error curve as the final stability error. Lower value means higher stability in a video. These three components are presented in the following.

\subsubsection{Fragment Error}
In fragment error, we evaluate the integrity of the detections along a trajectory. Particularly, the results of a stable detector should be consistent (always report as a target or always not). It should not frequently change its status throughout the trajectory. Formally, Let $N$ be the total number of trajectories in a video sequence. $t_k$ is the total length of the $k^{th}$ trajectory, and $f_k$ is the number of status change. One status change is defined as the scenario that the object is detected in previous frame but missed in current frame and vice versa. Then the fragment error is defined as:
\begin{equation}
E_F = \frac{1}{N}\sum_{k=1}^N\frac{f_k}{t_k - 1}.
\label{eq:fragment}
\end{equation}
As a special case, we define fragment error of a trajectory with length one to be 0. The fragment error is minimized when the detector can always localized the object accurately along the ground-truth trajectory or never been, while it is maximized when the object is detected alternately. It is also noteworthy that there is also one metric with same name~\cite{fragment} in the evaluation of MOT, however there is a key difference: Our metric is normalized by the trajectory length, while the one in MOT does not. This makes the metric comparable across different videos with different numbers of trajectories.


\subsubsection{Center Position Error}
\label{sec:centererr}
In center position error, we evaluate the stability of the center positions of the detections along a trajectory. This metric is illustrated in Fig.~\ref{fig:example}(c). A good detector should keep the centers of its outputs stable, instead of randomly jittering around the centers of the ground-truths. We evaluate the change of center position in both horizontal and vertical directions. For the predicted bounding box in the $f^{th}$ frame of trajectory $k$, we define it as $B_p^{k,f} = (x_p^{k,f}, y_p^{k,f}, w_p^{k,f}, h_p^{k,f})$ which is the center of horizontal axis, vertical axis, width and height. Similarly, the corresponding ground-truth is $B_{g}^{k,f} = (x_g^{k,f}, y_g^{k,f}, w_g^{k,f}, h_g^{k,f})$. Then the center position error is defined in Eq.(\ref{eq:centererr}), which is the average of the standard deviations of center positions in all trajectories.

\begin{equation}
\begin{aligned}
        e_x^{k,f} &= \frac{x_p^{k,f} - x_g^{k,f}} {w_g^{k,f}}, \quad \sigma_x^k = \text{std}(\e_x^k), \\ 
        e_y^{k,f} &= \frac{y_p^{k,f} - y_g^{k,f}} {h_g^{k,f}}, \quad \sigma_y^k = \text{std}(\e_y^k), \\
         E_C &= \frac{1}{N}\sum_{k=1}^N (\sigma_x^k + \sigma_y^k).
         \label{eq:centererr}
\end{aligned}
\end{equation}

It should be noted that the center position error only evaluates the variance of the normalized center deviation instead of its bias. The underlying reason is that the bias has been implicitly considered in the accuracy metric. In other words, larger bias will always result in lower accuracy. Also, as a consequence of this definition, if the detections in one trajectory consistently biased towards one direction, we will not penalize them. Though it may be wired at first sight, this is just the key that distinguishes accuracy and stability: They are bad detections in terms of accuracy, but good detections in terms of stability.

\subsubsection{Scale and Ratio Error}
Following the same spirit of center position error, we evaluate the stability of scale and aspect ratio of the detections along a trajectory. We demonstrate the idea in Fig.~\ref{fig:example}(d). Specifically, we use square root of the area ratio to represent the scale deviation, and define the ratio of two aspect ratios as aspect ratio deviation. The reason we apply square root to area ratio is that we need to keep the magnitude of each type of deviation same. At last, the final result is defined as the average of standard deviations of scale and ratio among all the trajectories. Formally, we have:

\begin{equation}
\begin{aligned}
        e_s^{k,f} &= \sqrt{\frac {w_p^{k,f} h_p^{k,f}} {w_g^{k,f} h_g^{k,f}}}, \quad &\sigma_s^k  = \text{std}(\e_s^k), \\ 
        e_r^{k,f} &= (\frac {w_p^{k,f}}{h_p^{k,f}}) / (\frac{w_g^{k,f}}{h_g^{k,f}}), \quad &\sigma_r^k  = \text{std}(\e_r^k), \\
        E_R &= \frac{1}{N}\sum_{k=1}^N(\sigma_s^k + \sigma_r^k).
        \label{eq:ratioerr}
\end{aligned}
\end{equation}

Same as the center position error, we also focus on the variance instead of the bias of the scale and ratio deviation in this metric. Consequently, if the detections are consistently larger or smaller than ground-truths, they will not be penalized.

\section{Relationships with Existing Metrics}
In this section, we discuss the connections and differences between our proposed metric and several commonly used metrics in VID and MOT.
\subsection{Relationship with CLEAR MOT}
\label{sec:mota}
The evaluation of MOT also emphasizes more on association in videos. The commonly used evaluation metrics in MOT are Multi-Object Tracking Accuracy (MOTA) and Multi-Object Tracking Precision (MOTP)~\cite{mota}. First, MOTP only considers the ability of localizing objects. It does not consider any trajectory related quality. Second, MOTA consists of three terms: false positive, false negative and identity switch of detected objects. The first two terms consider the detection accuracy in still images, while the last one considers the association accuracy in a trajectory. Our stability is built on the assumption that the association is already given (either output by the algorithm or find best matching with ground-truth), and assesses the stability based on that association.  Thus our stability metric captures a different aspect with the identity switch error in MOTA. 
\subsection{Relationship with tracklet IOU}

The subtask proposed in Imagenet ILSVRC 2016\footnote{\url{http://image-net.org/challenges/talks/2016/ILSVRC2016_10_09_vid.pdf}} also considers the temporal association. This new metric utilizes the output trajectory IDs as well as detection scores. It first sorts the trajectories by the mean detection scores, and then defines the tracklet IoU as the number of successfully detected frames over the number of the union of the output and ground-truth frames. Next, the mAP for each class is calculated based on this tracklet IoU. Although this novel metric considers both detection and association, it only focuses on temporal consistency. The stability along trajectories is still ignored. 

\section{Validation Setup}
In this section, we will introduce our datasets and basic models used in the experiments.

\subsection{Dataset}
Due to the numerous experiments needed in our validation, we choose two middle size datasets with ground-truth associations annotated. In particular, we use the MOT challenge datasets~\cite{MOT} and KITTI datasets~\cite{KITTI}. For the MOT challenge dataset, the annotations for testing set are not available, so we combine two years' MOT challenge training datasets for experiments. We use MOT 2016\footnote{\url{https://motchallenge.net/data/MOT16/}} training set for training, and MOT 2015\footnote{\url{https://motchallenge.net/results/2D_MOT_2015/}} training set for testing. We also exclude the overlapped videos for testing. For KITTI dataset, we evenly split them into training set and testing set. Note that we only use car category in KITTI dataset due to the reason that the number of annotated bounding boxes for pedestrian category is too small to train a stable detector, especially for deep learning based method.

\subsection{Basic Models}
Currently there is no end-to-end pipeline for VID. Most of them improve upon existing object detectors for still images. Therefore, in our experiment, we choose the following base detectors for validation: The first one is Aggregated Channel Feature (ACF). It is the representative work for non-deep learning method; while the second one is Faster RCNN\cite{fasterrcnn} which is the most popular CNN based object detector. We apply the VGG16 model~\cite{karen14vgg} which is pre-trained on the ImageNet classification task as our base CNN. For the post-processing of these two detectors, we both apply Non-Maximum Suppression (NMS) with threshold 0.5.

\section{Validation and Analysis}
In this section, we investigate several improvements for VID. We briefly divide them into two categories: The first type is to improve the aggregation of the output bounding boxes within a single frame. Specifically, we test the representative method weighted NMS~\cite{spyros2015object}. The second type is to utilize the temporal context across frames. Here we benchmark two methods, \ie Motion Guided Propagation (MGP)~\cite{kai16tcnn} and object tracking~\cite{medianflow}. We conduct ablation analyses to investigate how each component affects the final performance. 

Some existing works cannot be benchmarked include: 1) Multi-context suppression~\cite{kai16tcnn}: This method utilize the correlations among different classes to suppress false positives. However, in the two datasets we use, only one class exists. 2) Rescoring~\cite{seqnms,kai16tcnn}: We use the implementations from the original authors, but cannot get similar satisfactory results as reported in their papers on our datasets. We owe the reason to that MOT datasets often contain crowded scene that makes the trajectory generation fail. We will tackle this issue in our future work. 

\begin{figure}[!hbt]
    \centering   
    \subfigure[ACF on MOT]{\includegraphics[width=0.48\linewidth]{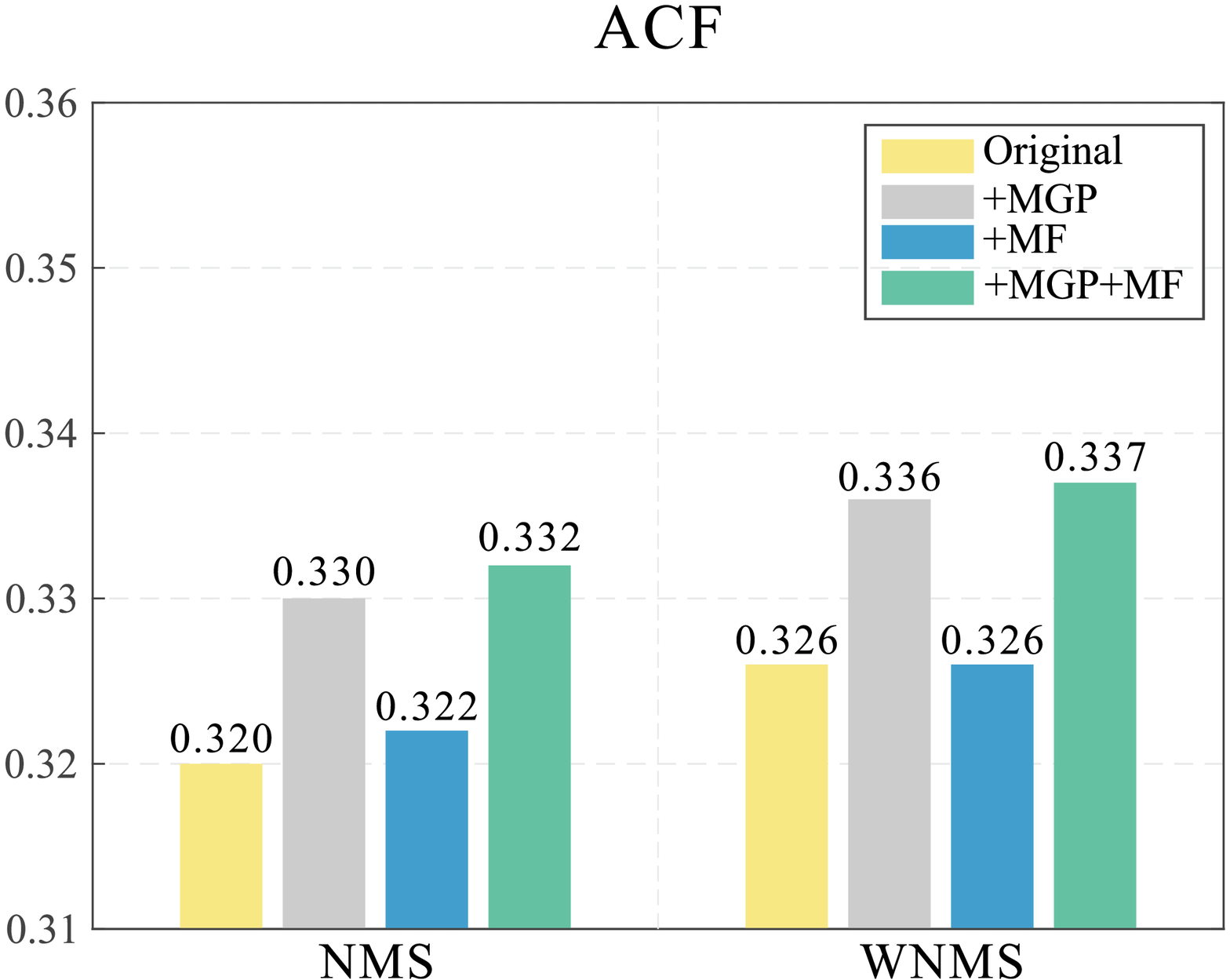}}
    \subfigure[Faster RCNN on MOT]{\includegraphics[width=0.48\linewidth]{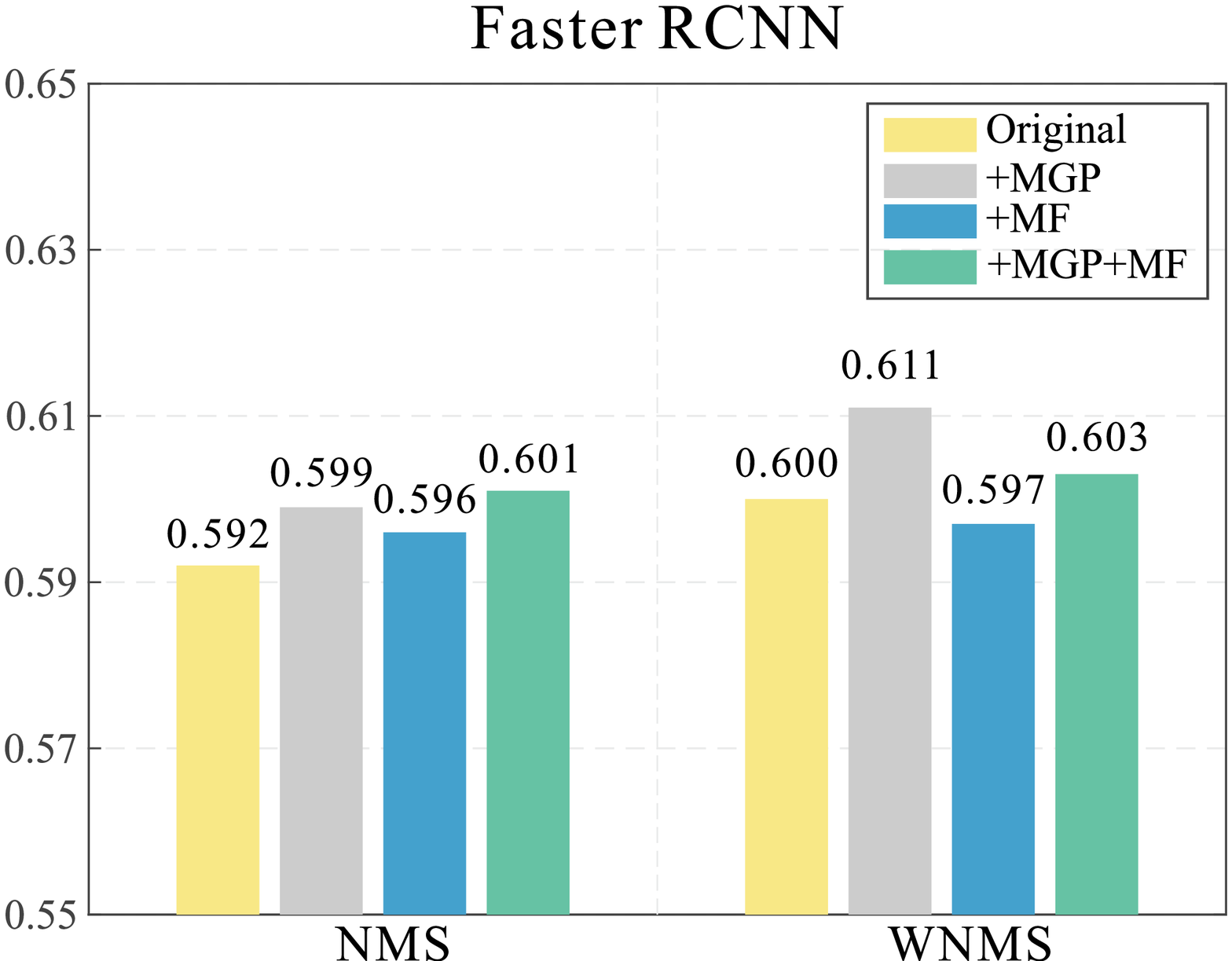}}
    \subfigure[ACF on KITTI]{\includegraphics[width=0.48\linewidth]{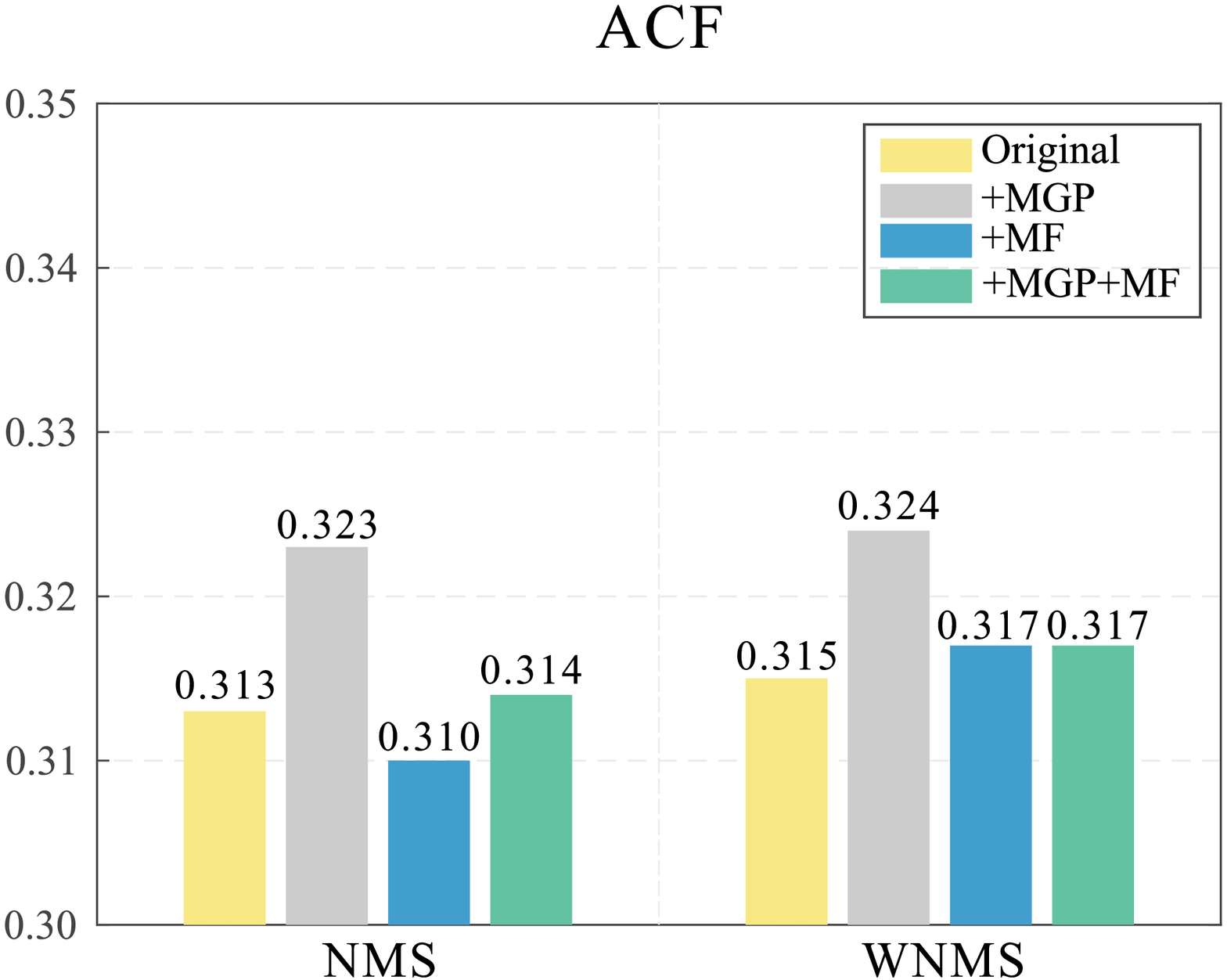}}
    \subfigure[Faster RCNN on KITTI]{\includegraphics[width=0.48\linewidth]{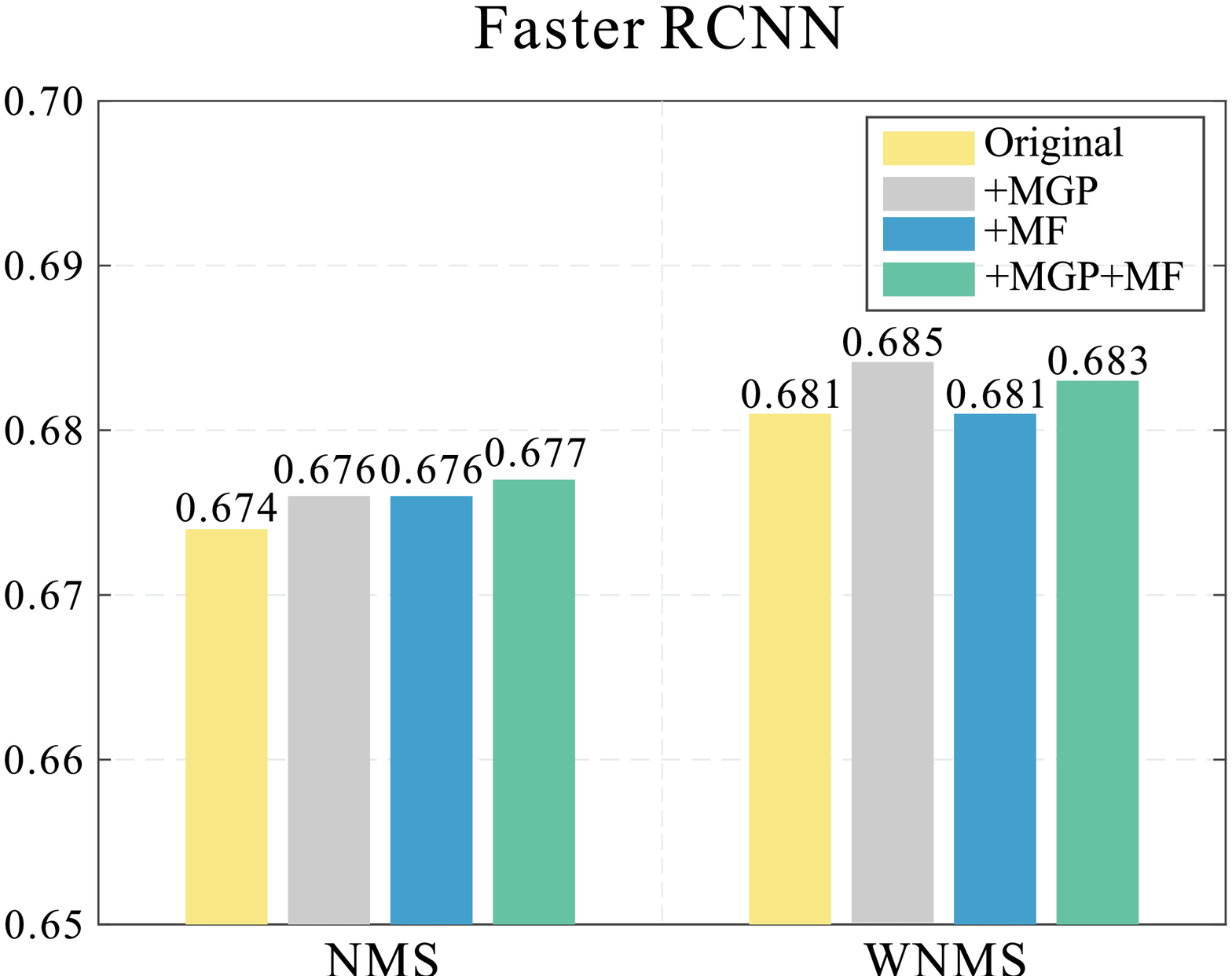}}
    \caption{Results of accuracy on MOT and KITTI detected by ACF and Faster RCNN.}
    \label{fig:accuracy}
\end{figure}

\begin{figure*}[!hbt]
    \centering   
    \subfigure[Stability Error]{\includegraphics[width=0.23\linewidth]{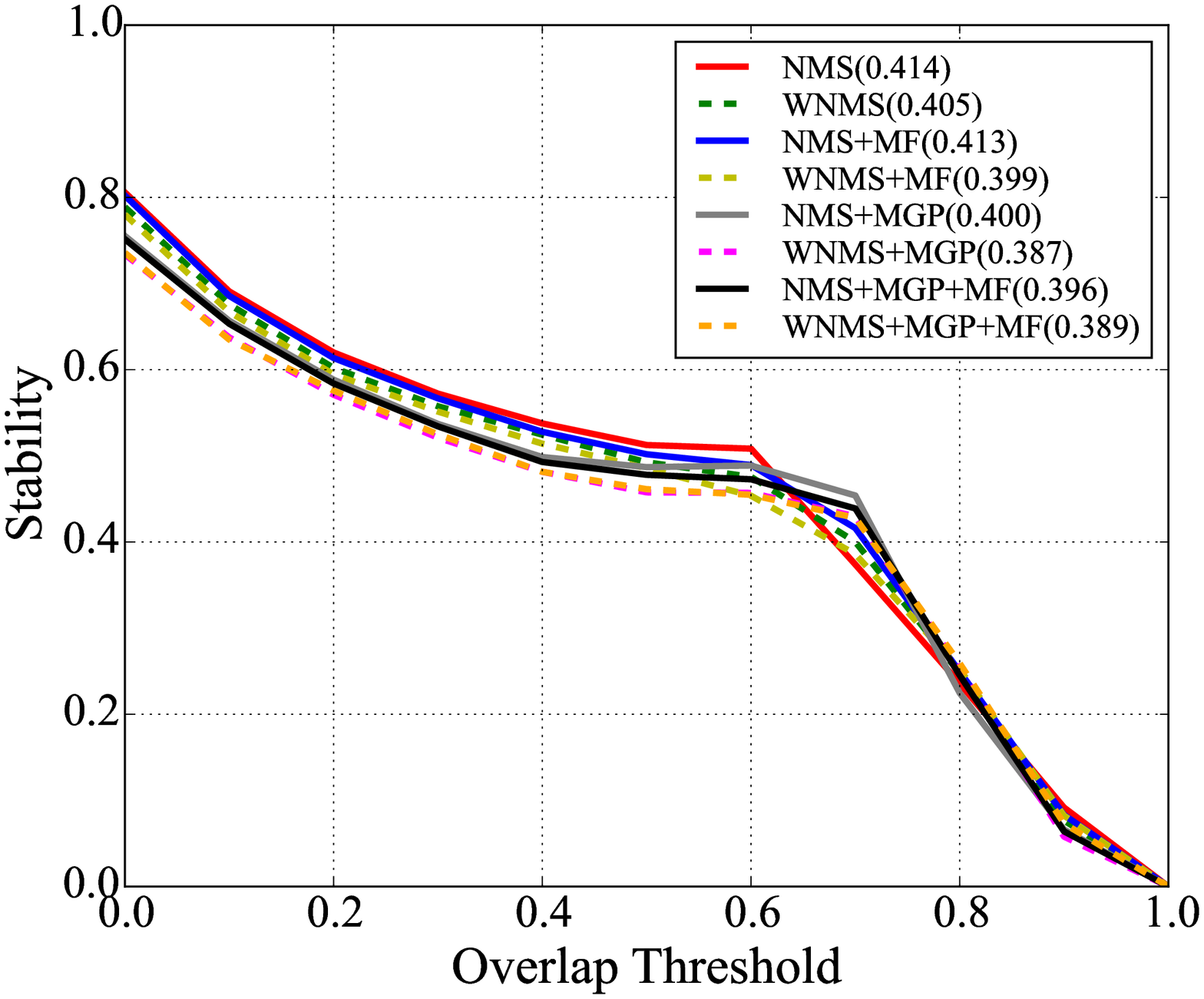}}
    \subfigure[Fragment Error]{\includegraphics[width=0.23\linewidth]{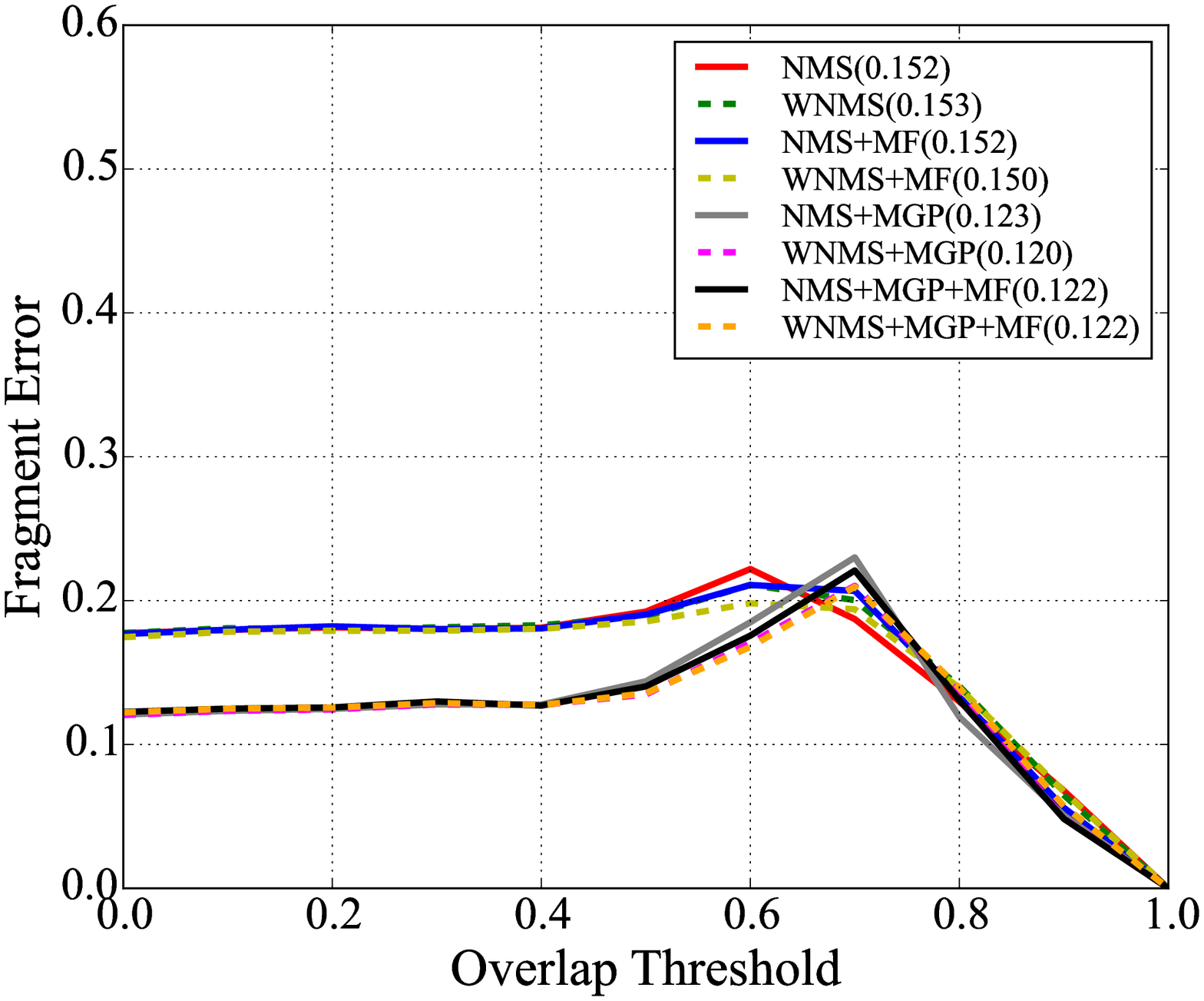}}
    \subfigure[Center Position Error]{\includegraphics[width=0.23\linewidth]{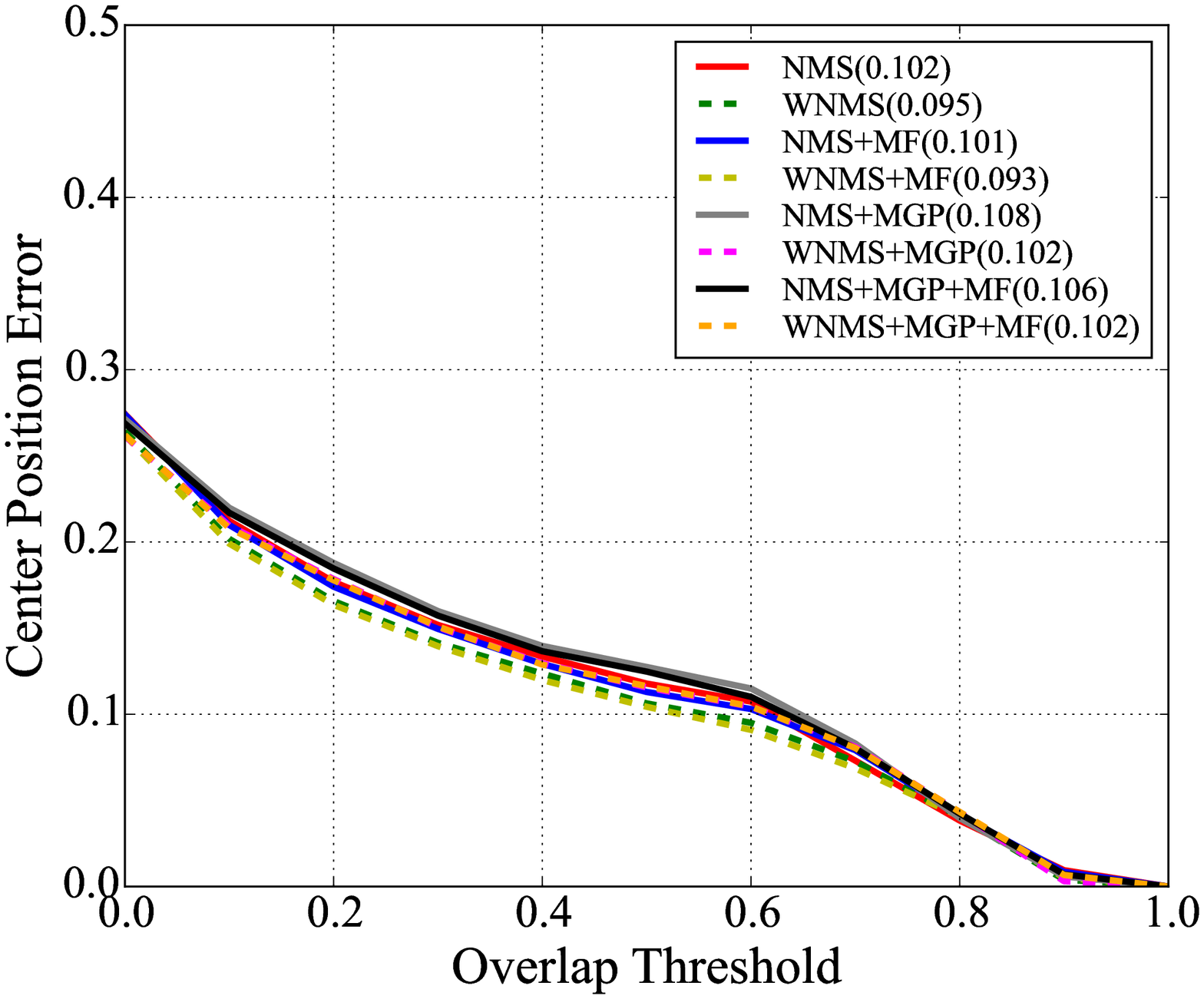}}
    \subfigure[Scale and Ratio Error]{\includegraphics[width=0.23\linewidth]{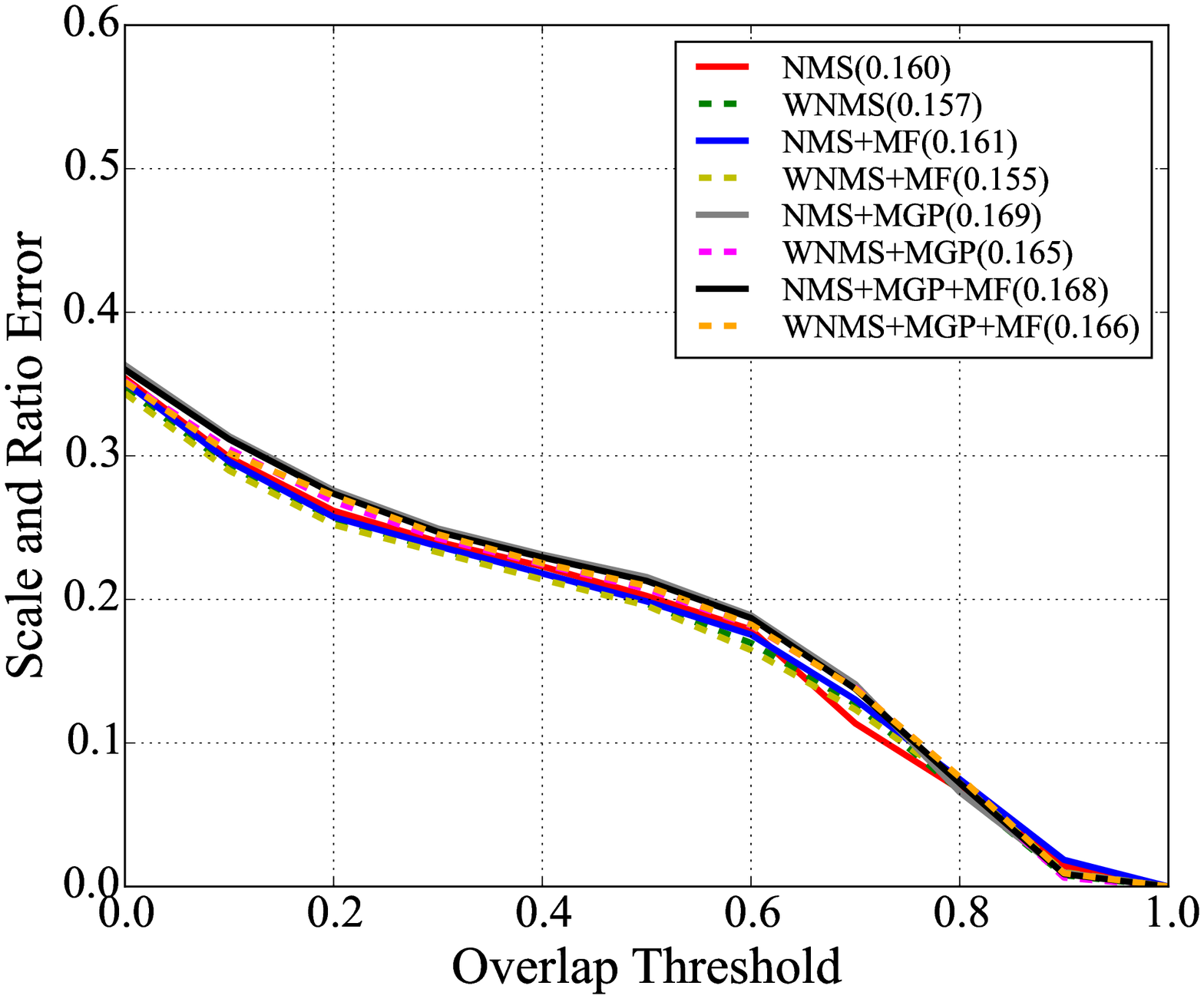}}\\
    \subfigure[Stability Error]{\includegraphics[width=0.23\linewidth]{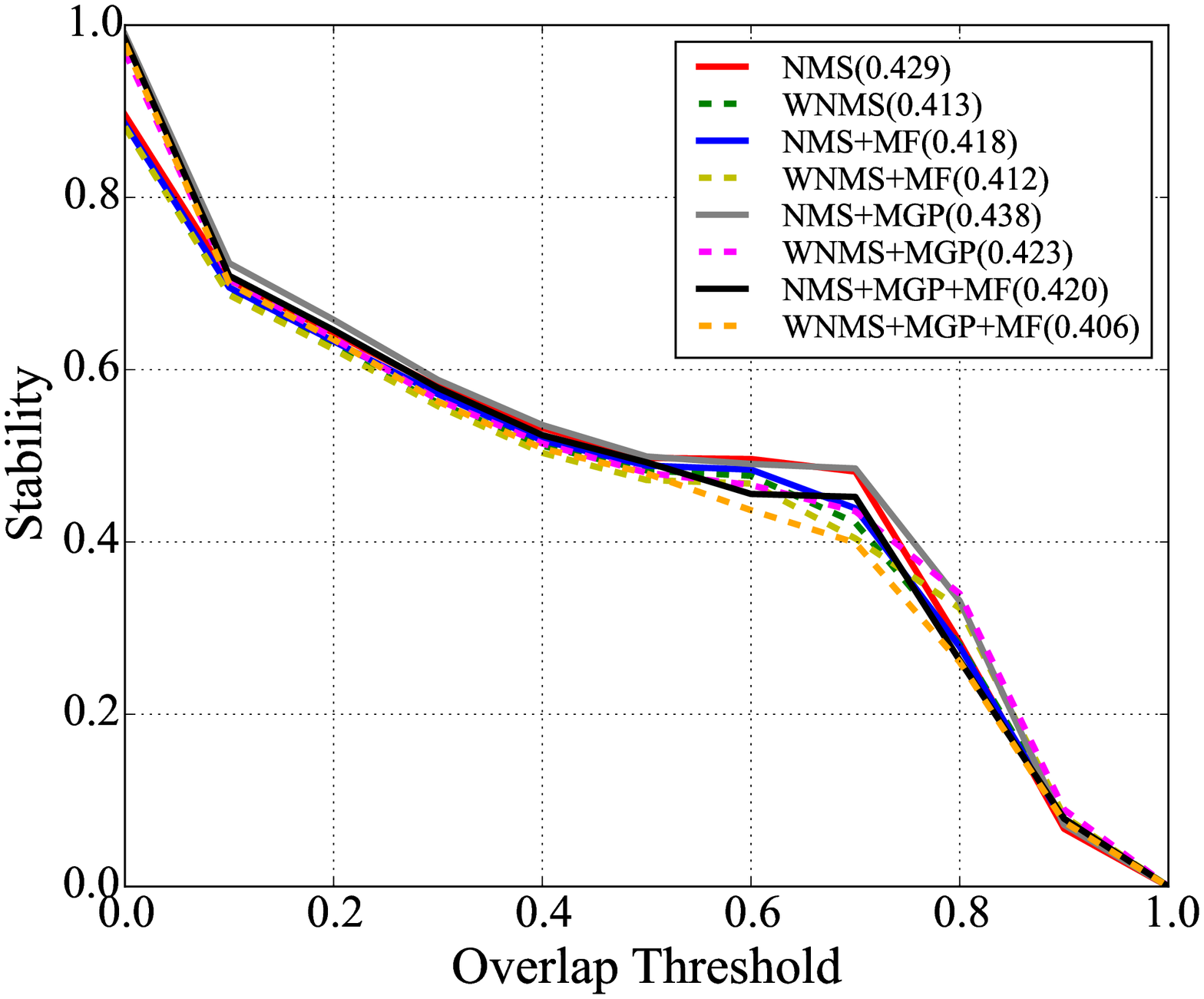}}
    \subfigure[Fragment Error]{\includegraphics[width=0.23\linewidth]{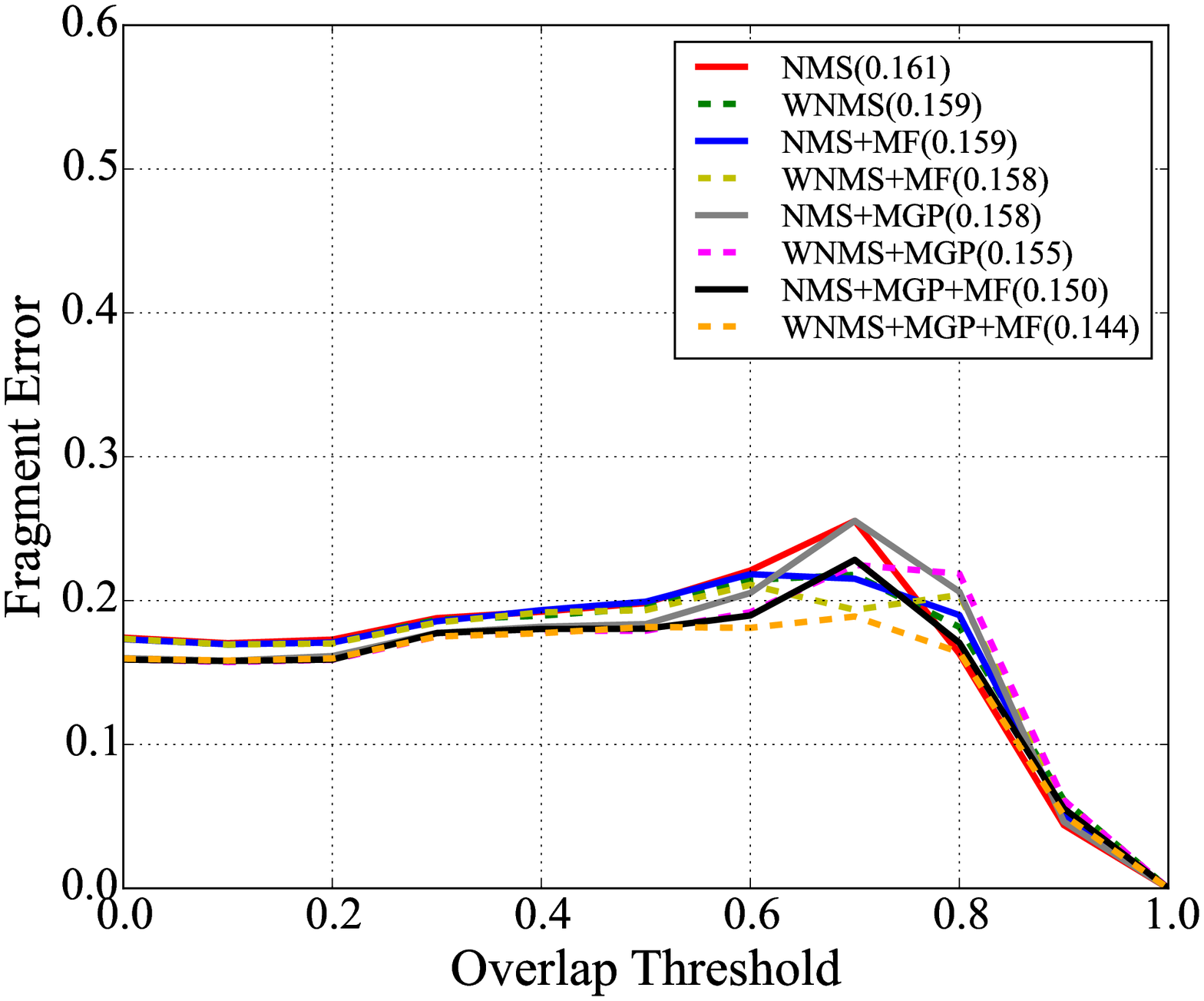}}
    \subfigure[Center Position Error]{\includegraphics[width=0.23\linewidth]{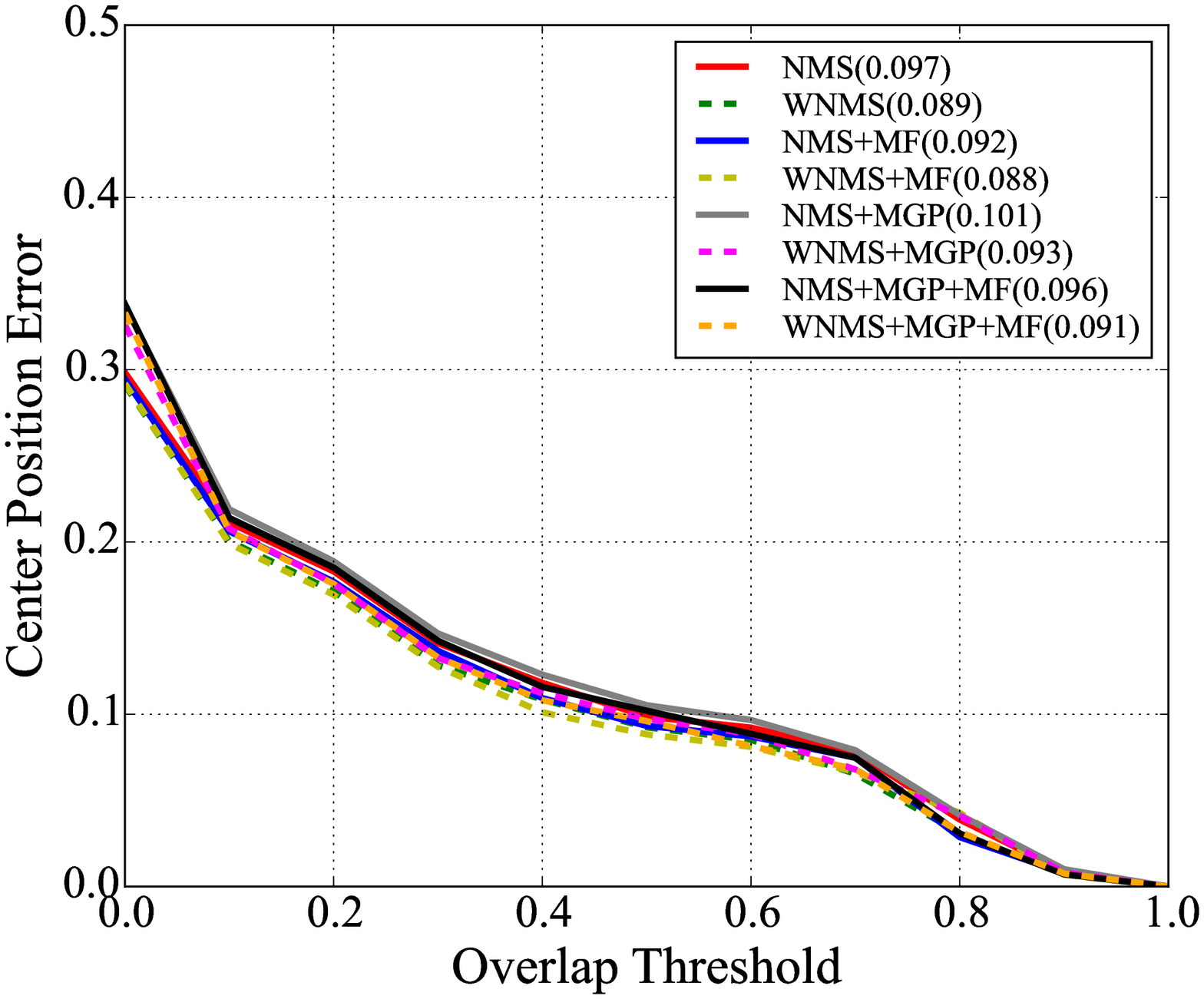}}
    \subfigure[Scale and Ratio Error]{\includegraphics[width=0.23\linewidth]{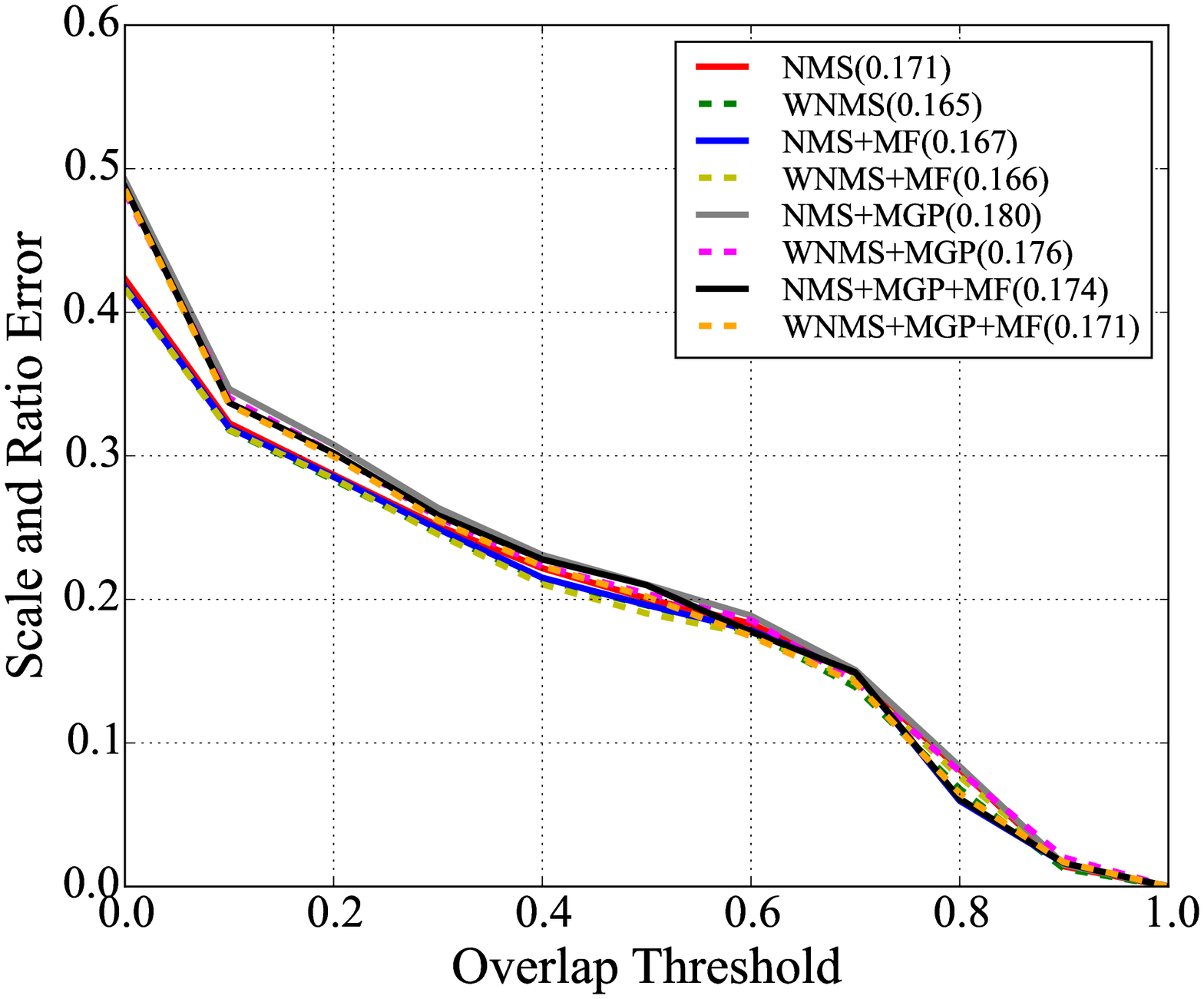}}
    \caption{Results of stability error of ACF. (a)-(d) on MOT, (e)-(h) on KITTI}
    \label{fig:motstacf}
\end{figure*}

\begin{figure*}[!htb]
    \centering   
    \subfigure[Stability Error]{\includegraphics[width=0.23\linewidth]{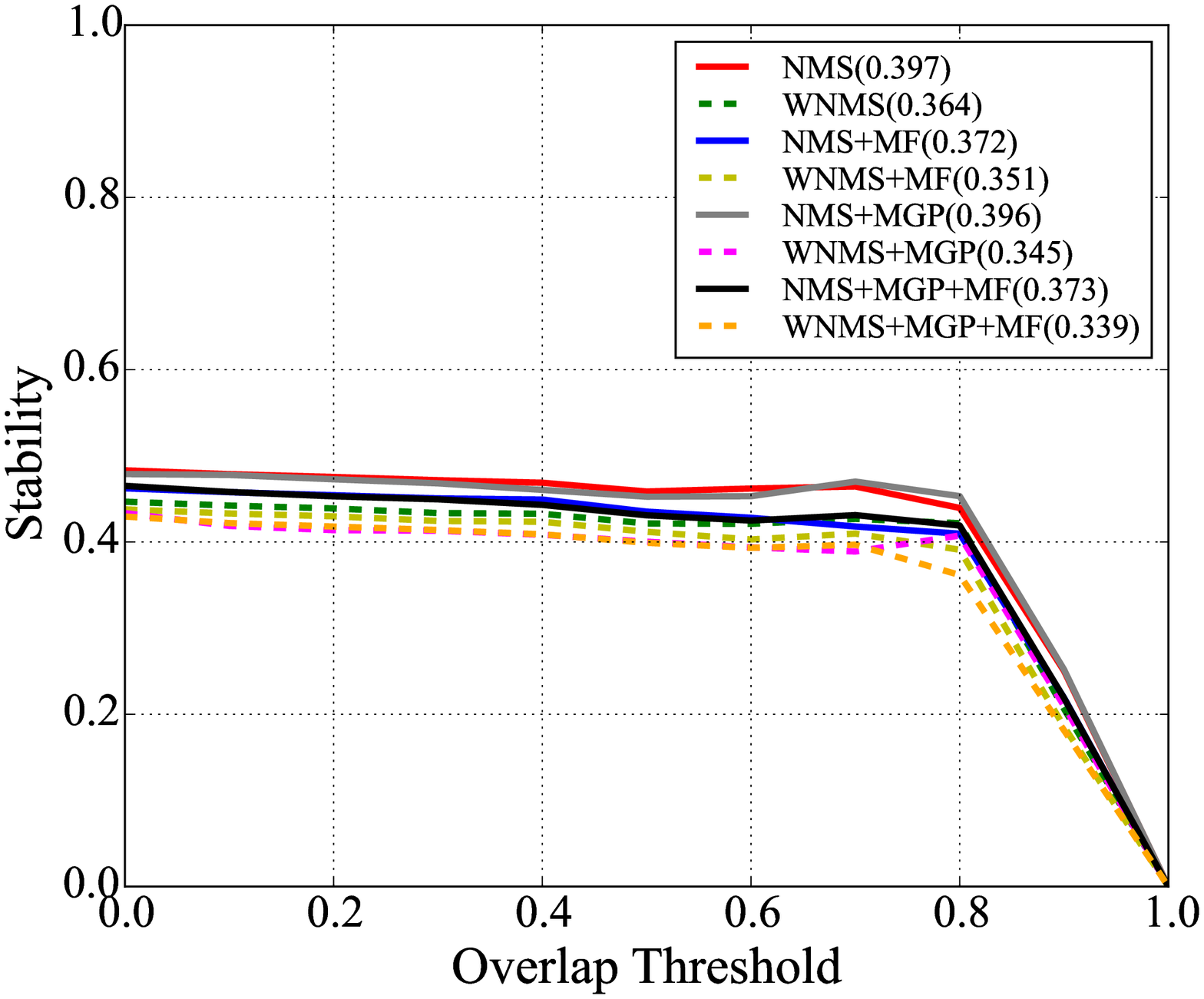}}
    \subfigure[Fragment Error]{\includegraphics[width=0.23\linewidth]{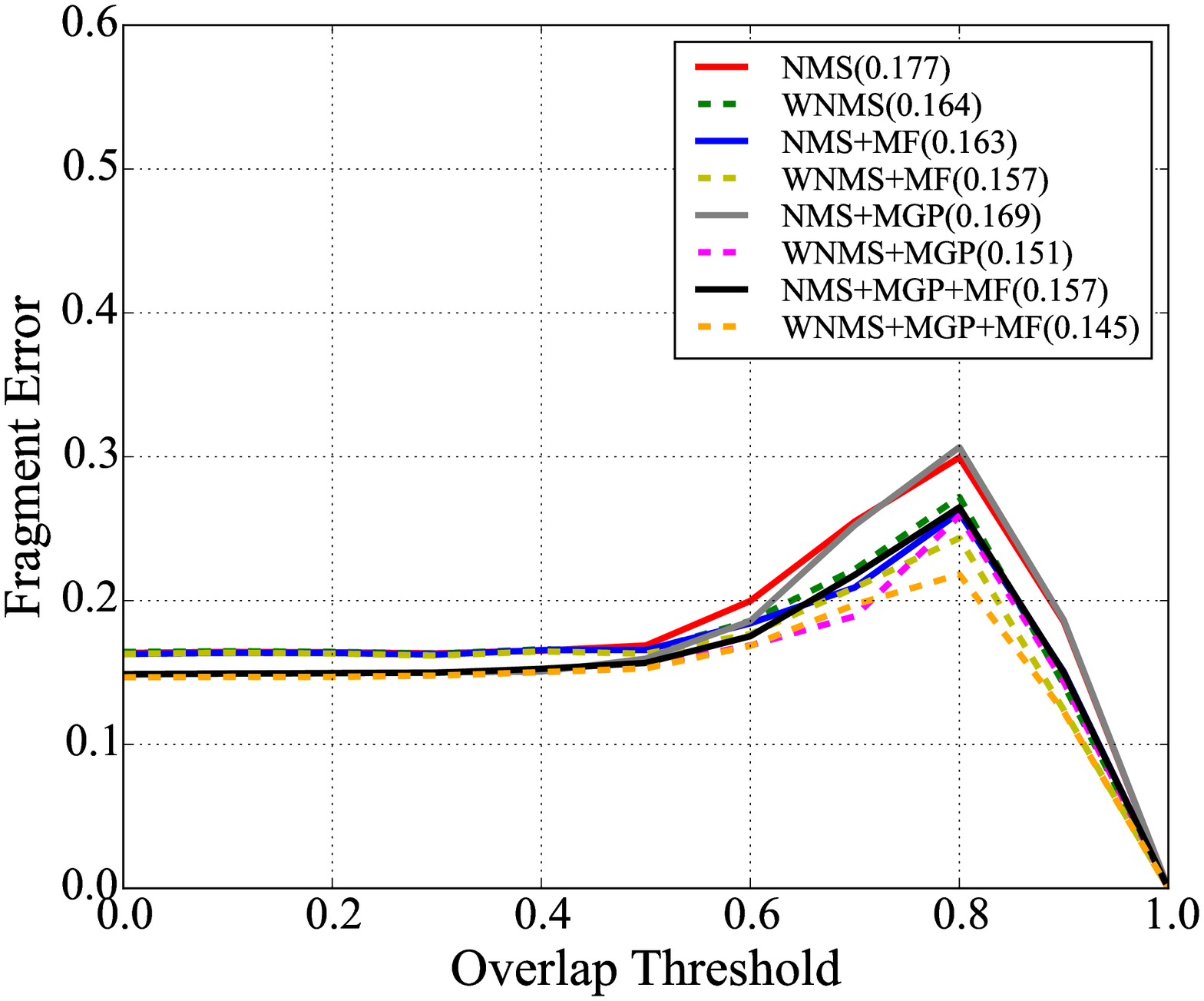}}
    \subfigure[Center Position Error]{\includegraphics[width=0.23\linewidth]{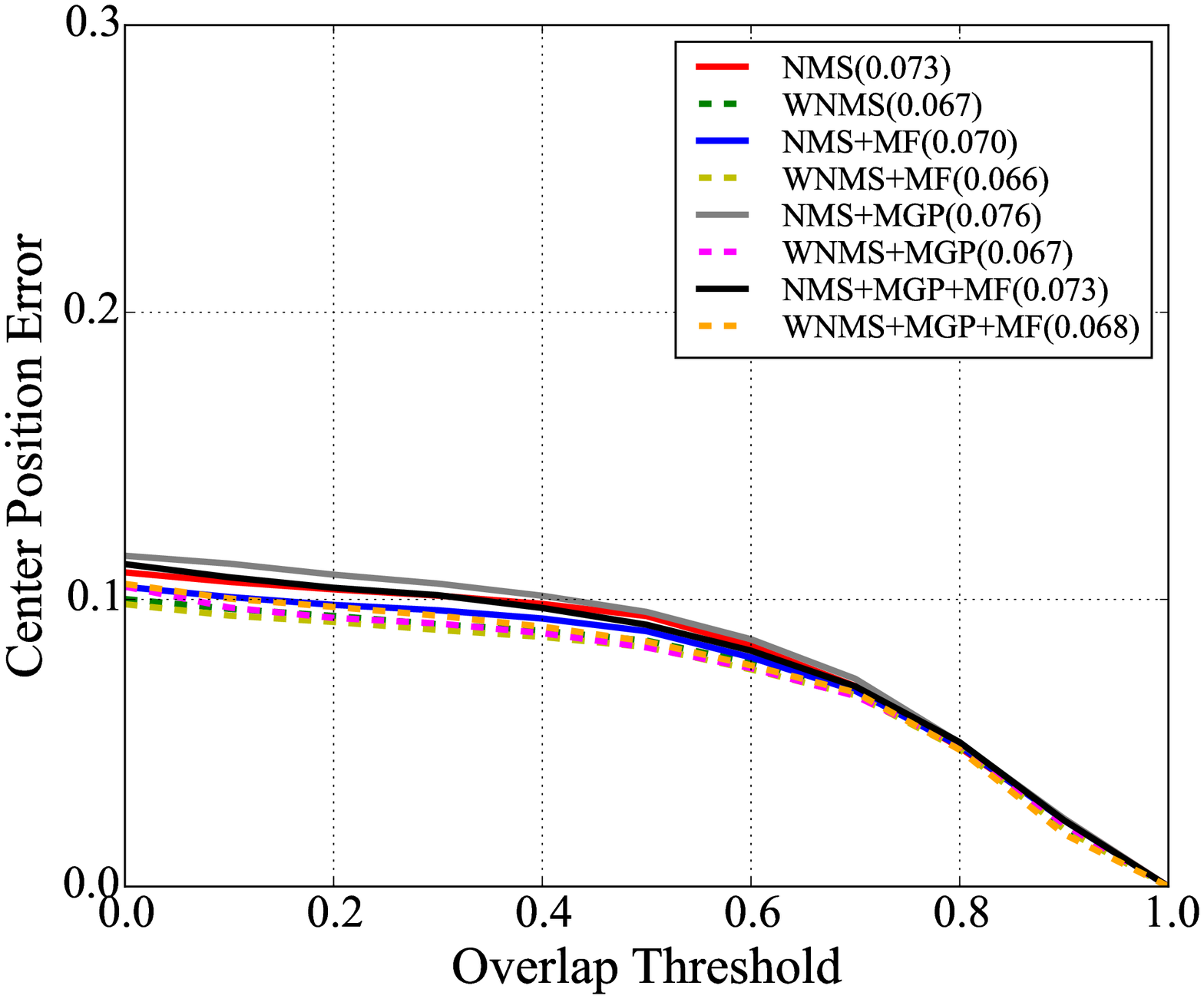}}
    \subfigure[Scale and Ratio Error]{\includegraphics[width=0.23\linewidth]{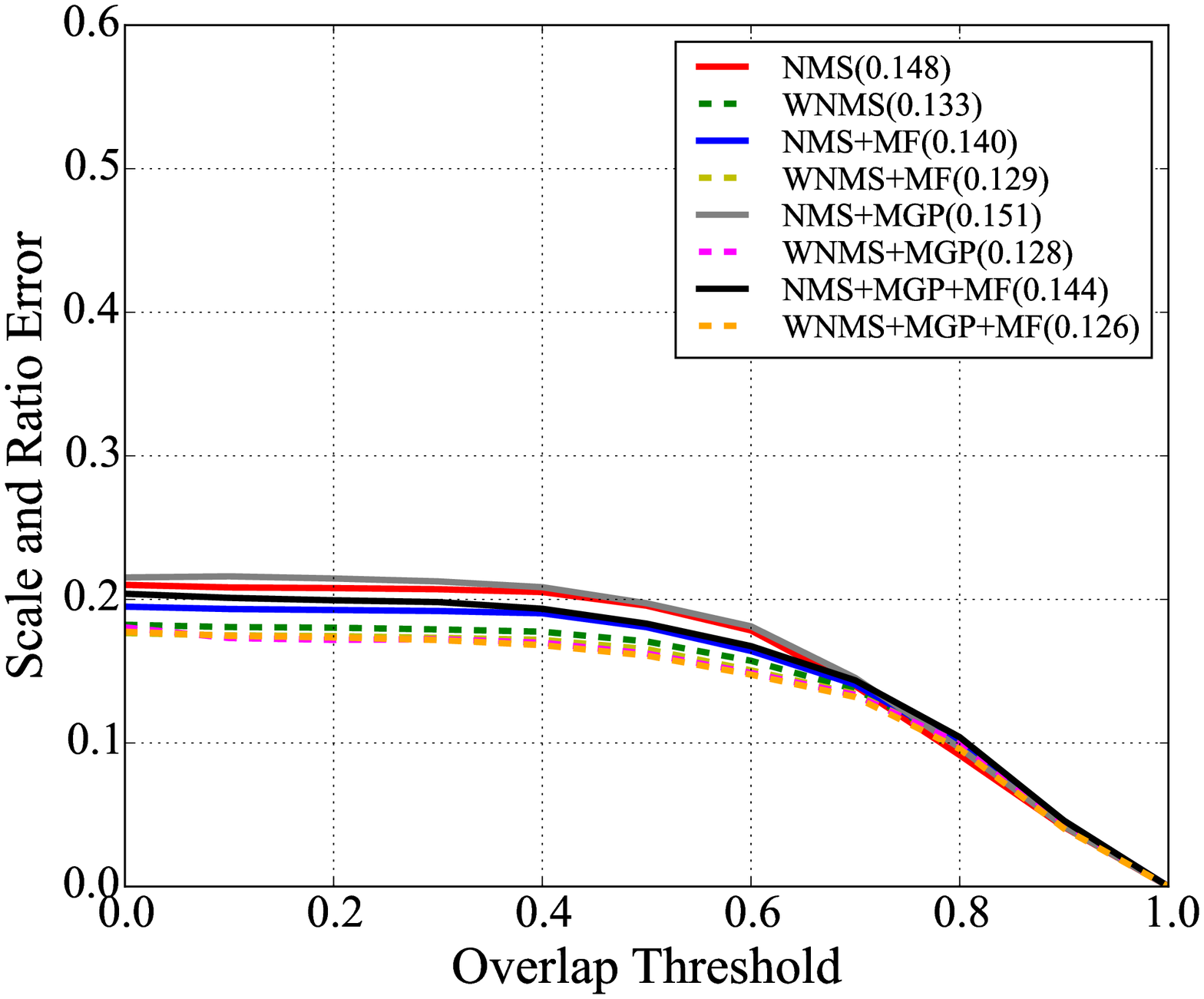}}\\
    \subfigure[Stability Error]{\includegraphics[width=0.23\linewidth]{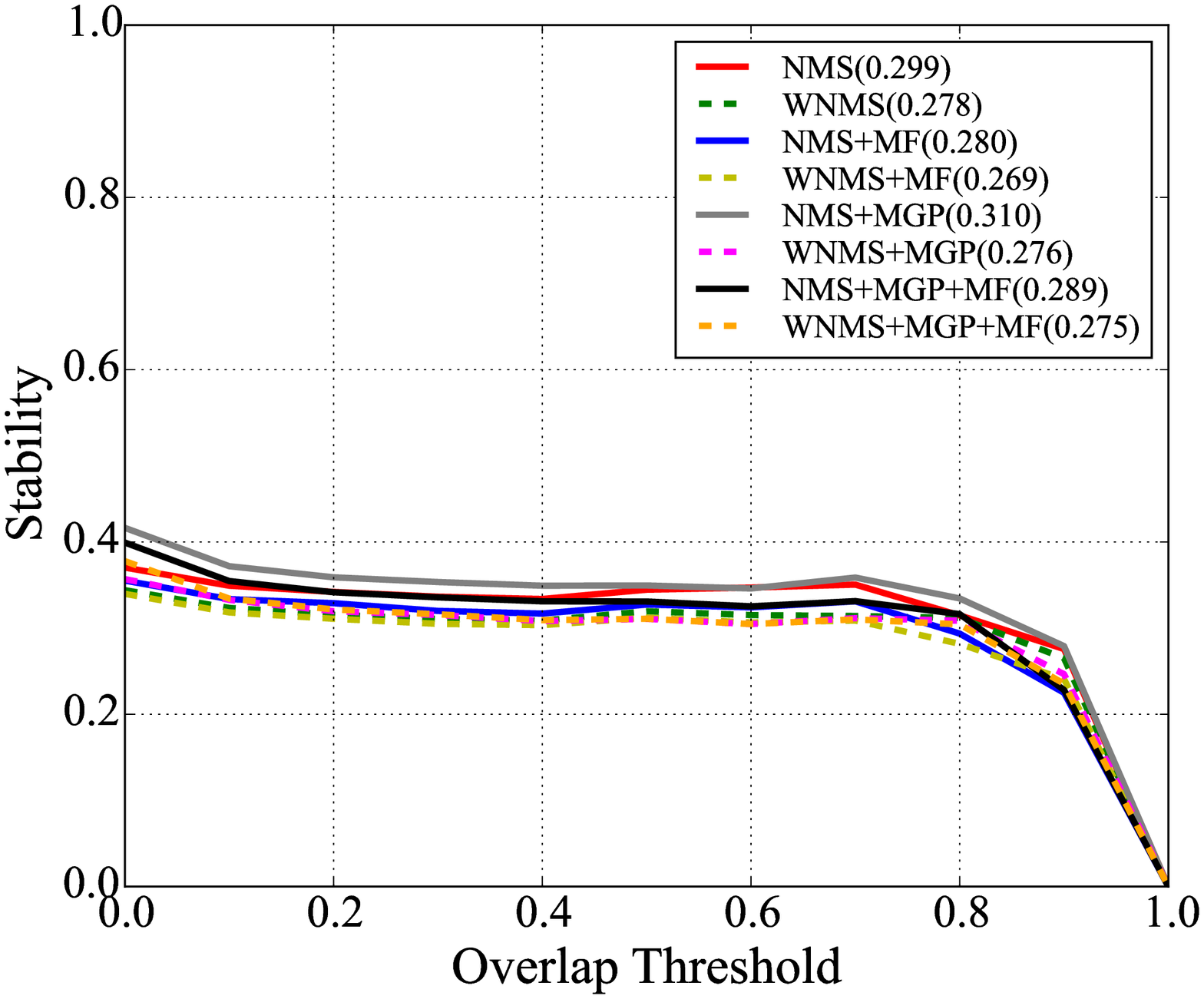}}
    \subfigure[Fragment Error]{\includegraphics[width=0.23\linewidth]{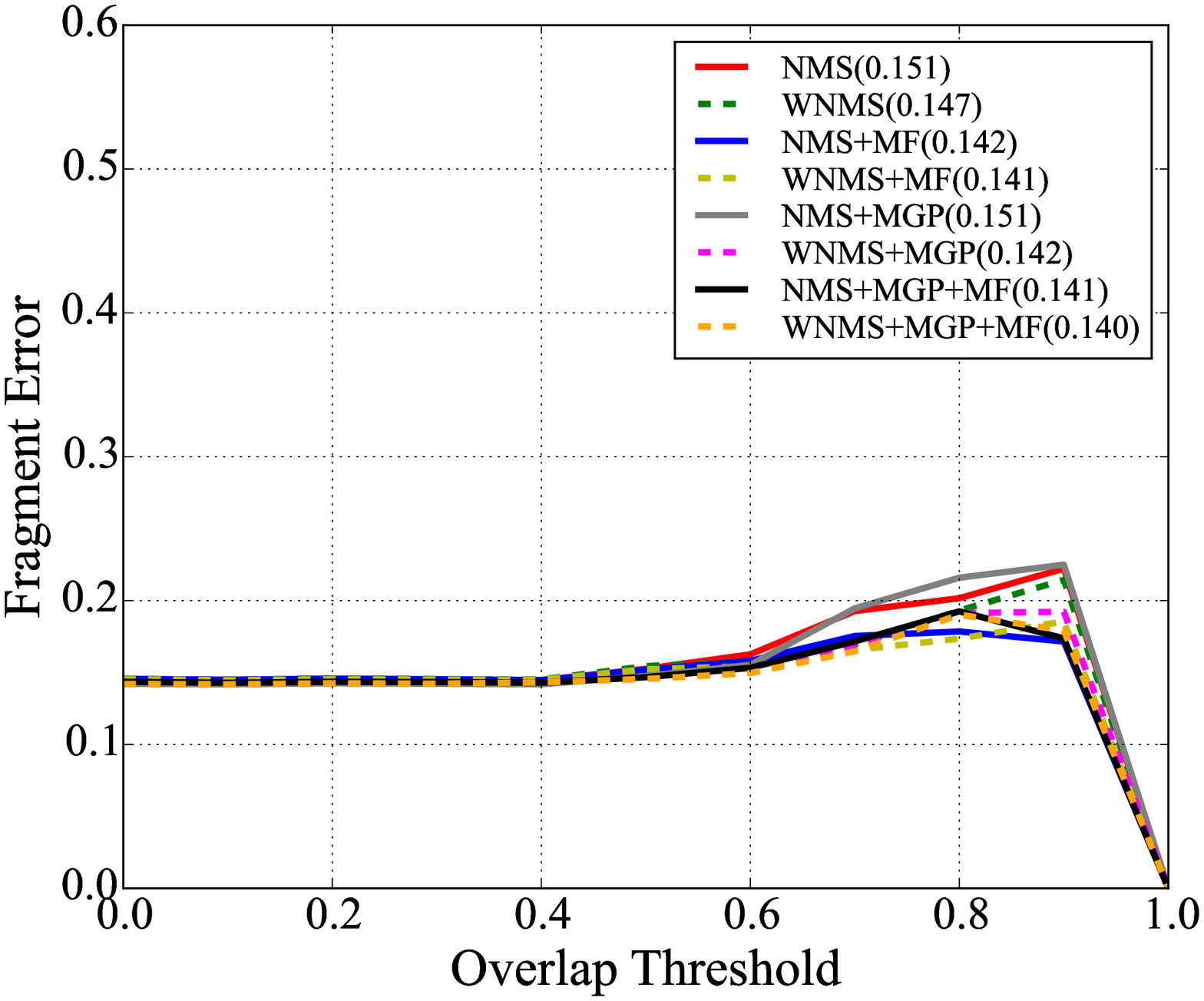}}
    \subfigure[Center Position Error]{\includegraphics[width=0.23\linewidth]{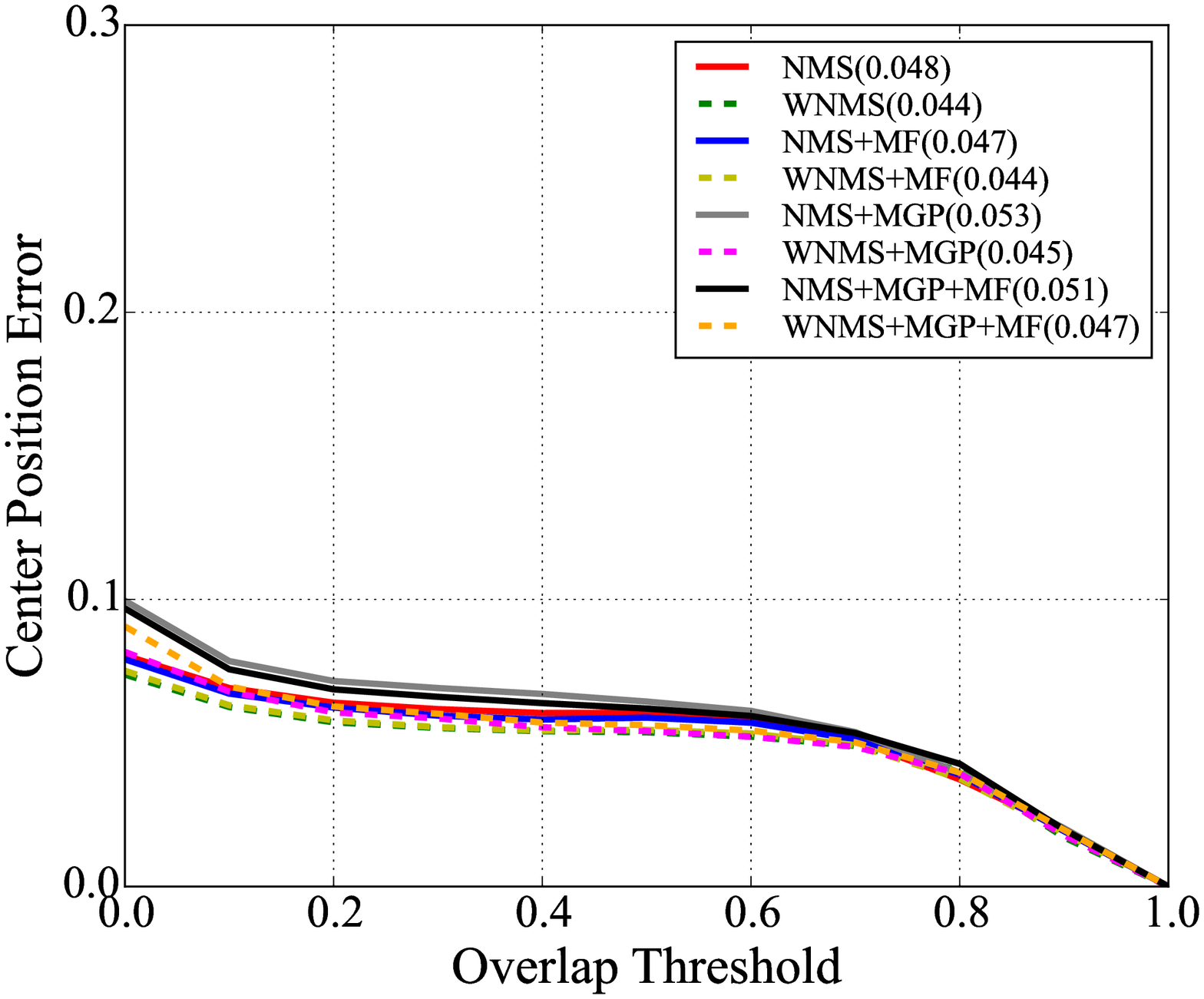}}
    \subfigure[Scale and Ratio Error]{\includegraphics[width=0.23\linewidth]{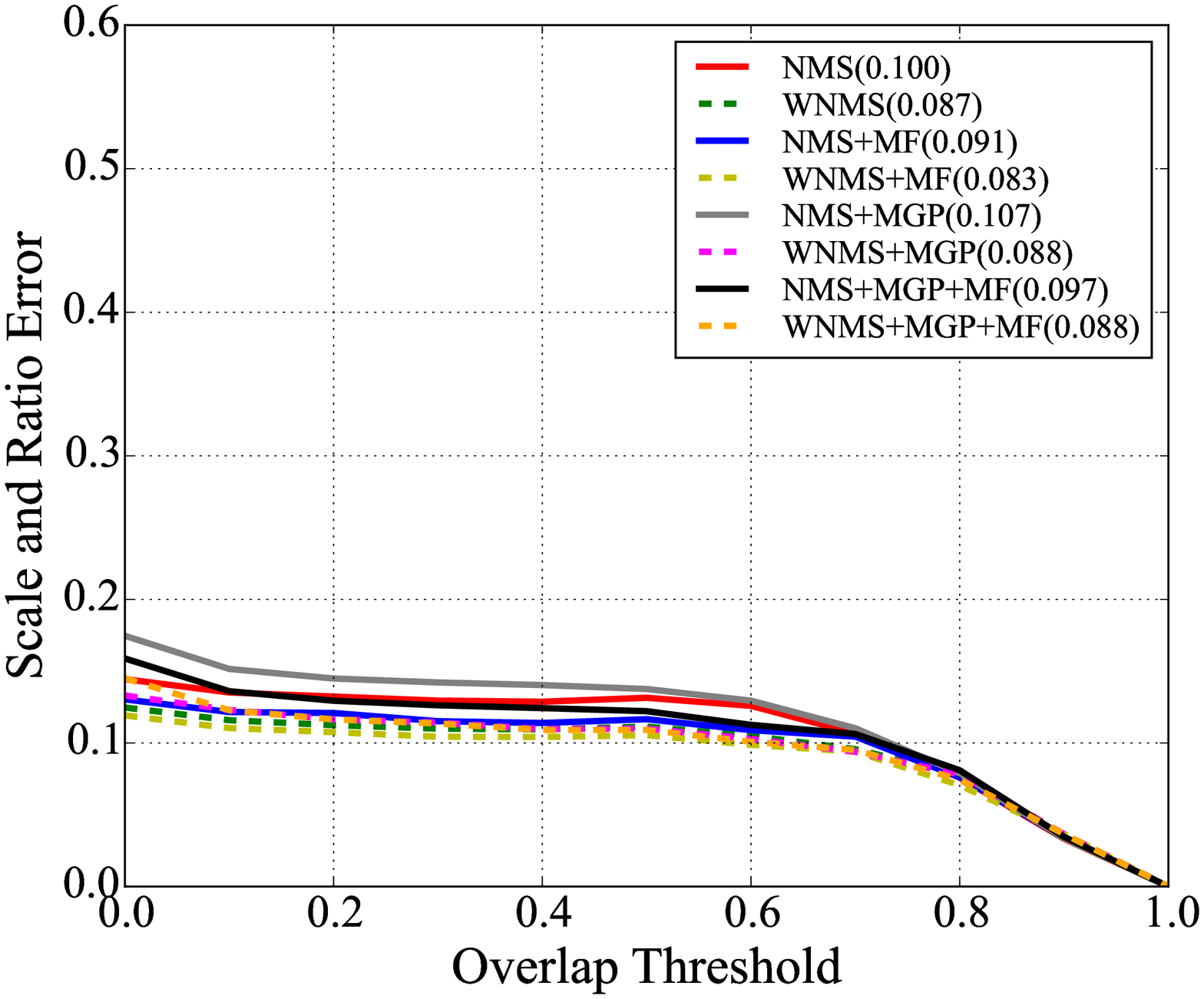}}
    \caption{Results of stability error of Faster RCNN. (a)-(d) on MOT, (e)-(h) on KITTI}
    \label{fig:motstfr}
\end{figure*}

All detection accuracy results are shown in Fig.~\ref{fig:accuracy} and detection stability results are shown in Fig.~\ref{fig:motstacf},~\ref{fig:motstfr}, respectively. \emph{We strongly recommend the readers watch the supplemental videos for intuitive comparisons.}\footnote{\url{https://tinyurl.com/lduu8y8}}

\subsection{Weighted NMS}
In a typical pipeline of object detection, after a classifier scores each proposal, usually an aggregation method is adopted to suppress the redundant bounding boxes. We argue that improper aggregation will significantly lower both accuracy and stability. The original NMS iterates between selecting the unsuppressed bounding box with highest score and suppressing all bounding boxes with an IOU higher than a given threshold with it. However, this method misses the valuable treasures from the suppressed bounding boxes. They still provide useful statistics about the objects. So in weighted NMS, rather than only keeping the bounding box with highest score, we weighted average it with all the suppressed bounding boxes by their scores. It was first proposed in~\cite{spyros2015object} to improve the mAP of still image detection. Interestingly, we find that it is also helpful to improve stability. In the sequel, we use WNMS for short. Note that there are also other advanced aggregation methods such as~\cite{liu2015box} which uses a learned function from data for aggregation. Due to limited space, we only benchmark the most representative one -- weighted NMS.

\paragraph{Result Analysis:} Weighed NMS consistently improves over NMS in both accuracy and stability. Notably, the marginal benefit even increases: The gap of these two methods for Faster RCNN is even larger than that of ACF. We owe the reason to the increased complexity of the detection system: A complicated system may reduce the bias, however it may not reduce, even increase the variance of the outputs. Weighted NMS effectively makes up this disadvantage by averaging over samples. Besides its effectiveness, it is also easy to implement and almost cost free.

\subsection{Motion Guided Propagation}
Due to the high correlation of adjacent frames, propagating detections to adjacent frames may help recover false negatives. Motion Guided Propagation (MGP) is proposed by this spirit in~\cite{kai16tcnn}. MGP takes the raw bounding boxes before aggregation, and then propagates them bidirectionally across adjacent frames using optical flow. The propagated bounding boxes are treated equally as other detections, and are used in the subsequent aggregation. Different from the original paper, we empirically find that adding a decay factor for the detection score for each propagation could improve the results. The decay factor is dataset dependent, and we tune it using  validation set. 


\paragraph{Result Analysis:} 
MGP consistently improves the accuracy for both ACF and Faster RCNN. In stability, MGP is especially helpful for fragment error. This is reasonable since MGP propagates detections with high confidence bidirectionally. It is not surprising that it helps to recover the false negatives. As for center error and scale and ratio error, the impact is uncertain for different datasets.


\subsection{Object Tracking}
We next investigate the use of object tracking to smooth the trajectory. In our implementation, we choose Median Flow (MF)~\cite{medianflow}. It is proven to be an efficient and effective short term tracking method. The merit of it lies in that it can detect self failure reliably by checking forward-backward error. Different from MGP, we apply MF after the bounding box aggregation phase (NMS or weighted NMS). Moreover, we only use it to smooth the detection bounding box, and do not alter the detection score or add new detection bounding boxes. Concretely, we start tracking with high confident detections (detection score higher than 0.8 in our experiments). If the tracker reports a reliable tracking result, we find the bounding box with highest IOU (should be at least higher than a given threshold, 0.5 in our case.) with it, and then average them as the final bounding boxes. For the detections without any associated tracking bounding box, we keep it unchanged. 

\paragraph{Result Analysis:} As expected, median flow is effective at stabilizing the detections, but has minor effect on accuracy. Even though median flow is one simple tracker and we only average the results, tracking is still beneficial to the performance. We believe that a more sophisticated tracker and better fusion method will further improve the results.

\subsection{Methods Combination}
\begin{figure*}[!htb]
    \centering   
\includegraphics[width=0.9\linewidth]{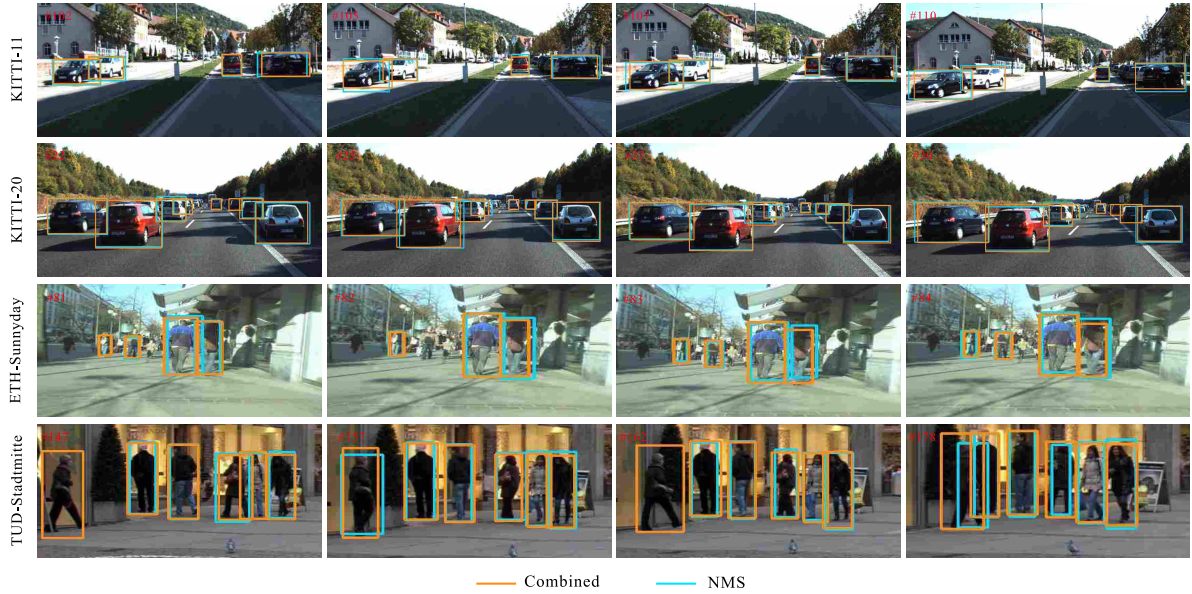}
\caption{Selected examples on MOT and KITTI using NMS and our best combination of methods. The base detector is Faster RCNN.}
\label{fig:cases}
\end{figure*}

Finally, we put together all these improvements. The best combination consistently improves over the baseline 1.0\texttildelow 2.0 mAUC in accuracy and 0.03\texttildelow0.06 (8\%\texttildelow15\% relative improvement) in stability. We further illustrate some visual results in Fig.~\ref{fig:cases}. It is easy to observe that the combined method localizes the objects of interest more accurately and consistently compared to the baseline, especially in the crowded and occluded scene.

\section{Metric Analysis}
In this section, we investigate the relationship between accuracy and stability in the proposed metric. Through the analysis, we justify that these two metrics are complementary and both necessary to assess the performance of a detector in VID.

\subsection{Correlation Analysis}
\begin{figure}[!hbt]
\centering
\begin{tabular}{@{\hspace{0mm}}c}
\includegraphics[width=0.6\linewidth]{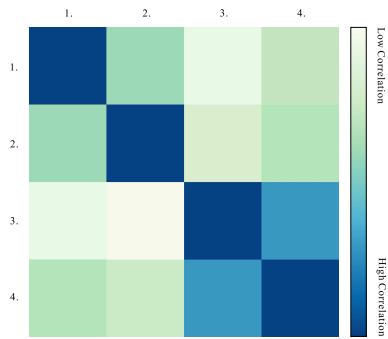}
\\
\end{tabular}
\caption{Correlation matrix for four measurements. 1: detection accuracy, 2: fragment error, 3: center position error, 4: scale and ratio error. Best viewed in color.}
\label{fig:corr_mat}
\end{figure}

Inspired by the correlation analysis of robustness and accuracy for single object tracking~\cite{vcehovin2016visual}, 
we also analyze the correlation between all the metrics including the accuracy and three types of stability. We run Faster RCNN and ACF with 8 different methods on both 7 video sequences of MOT and 9 video sequences of KITTI. As a result, we have 256 samples for each metric. We plot the absolute value of the correlation matrix in Fig.~\ref{fig:corr_mat}. For each cell in the figure, darker color denotes higher correlation between corresponding metrics.

It is easy to see that the accuracy metric has relatively low correlation with other three stability metrics. This is anticipated because they should characterize different aspects of quality in VID. Thus it is meaningful to measure both of them. Then we take a closer look at the three stability metrics. Among them, center position error and scale and ratio error are the most correlated, and fragment error is less correlated with them. This is reasonable since center position error and scale and ratio error both represent the spatial stability in one trajectory, while fragment error represents the temporal stability.

\subsection{Accuracy vs Stability}
\begin{figure}[!htb]
    \centering   
    \subfigure[ACF on MOT]{\includegraphics[width=0.48\linewidth]{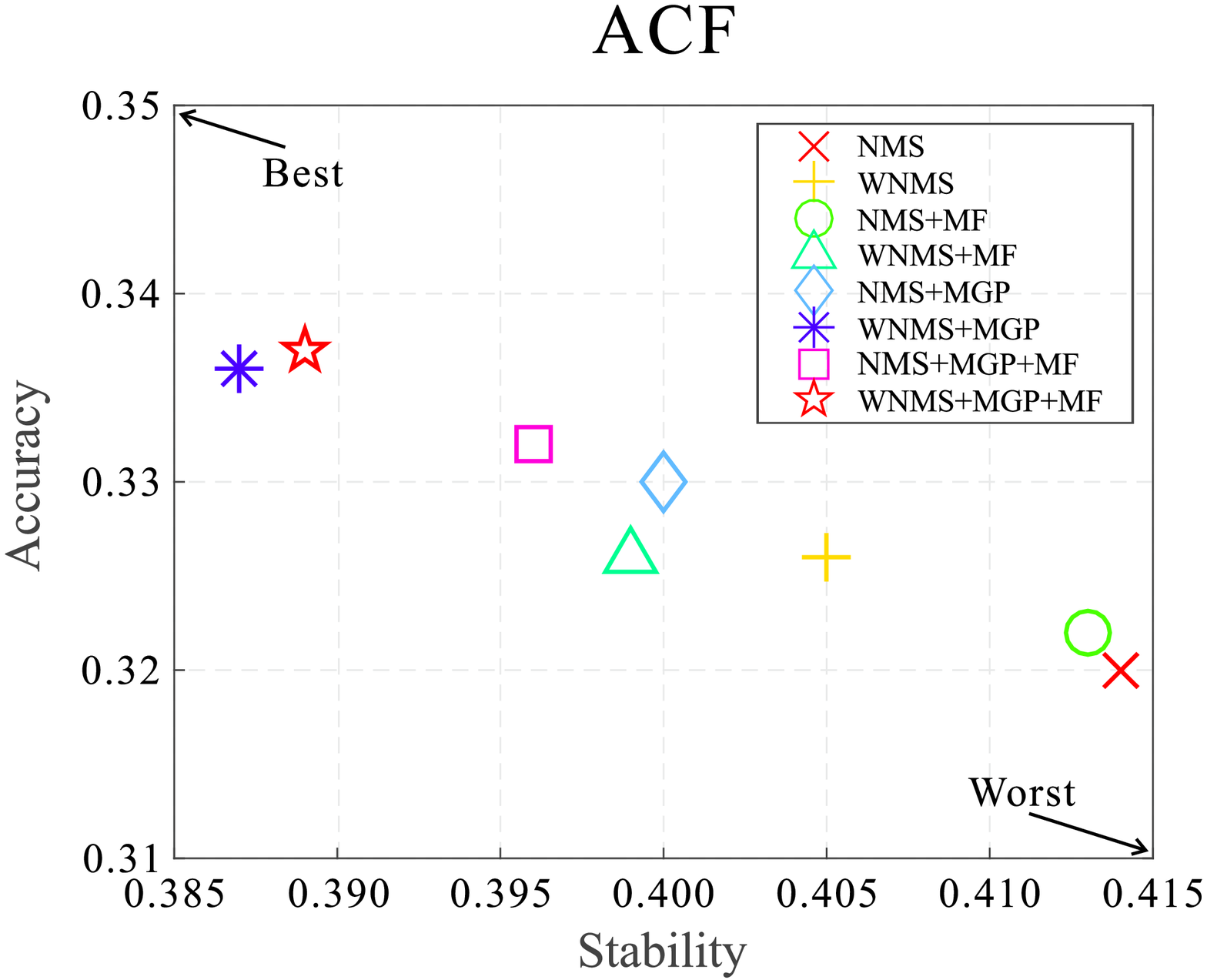}}
    \subfigure[Faster RCNN on MOT]{\includegraphics[width=0.48\linewidth]{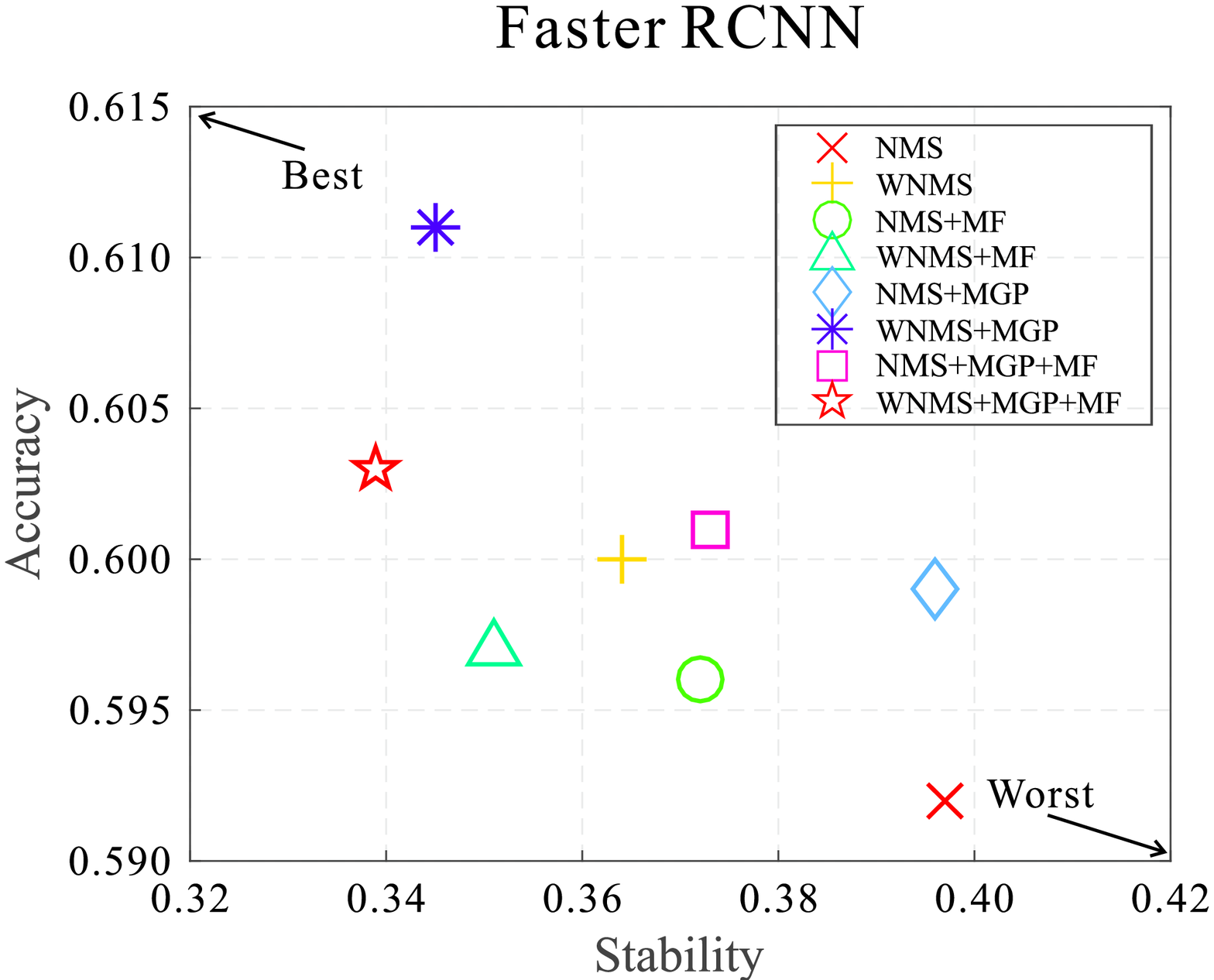}}
    \subfigure[ACF on KITTI]{\includegraphics[width=0.48\linewidth]{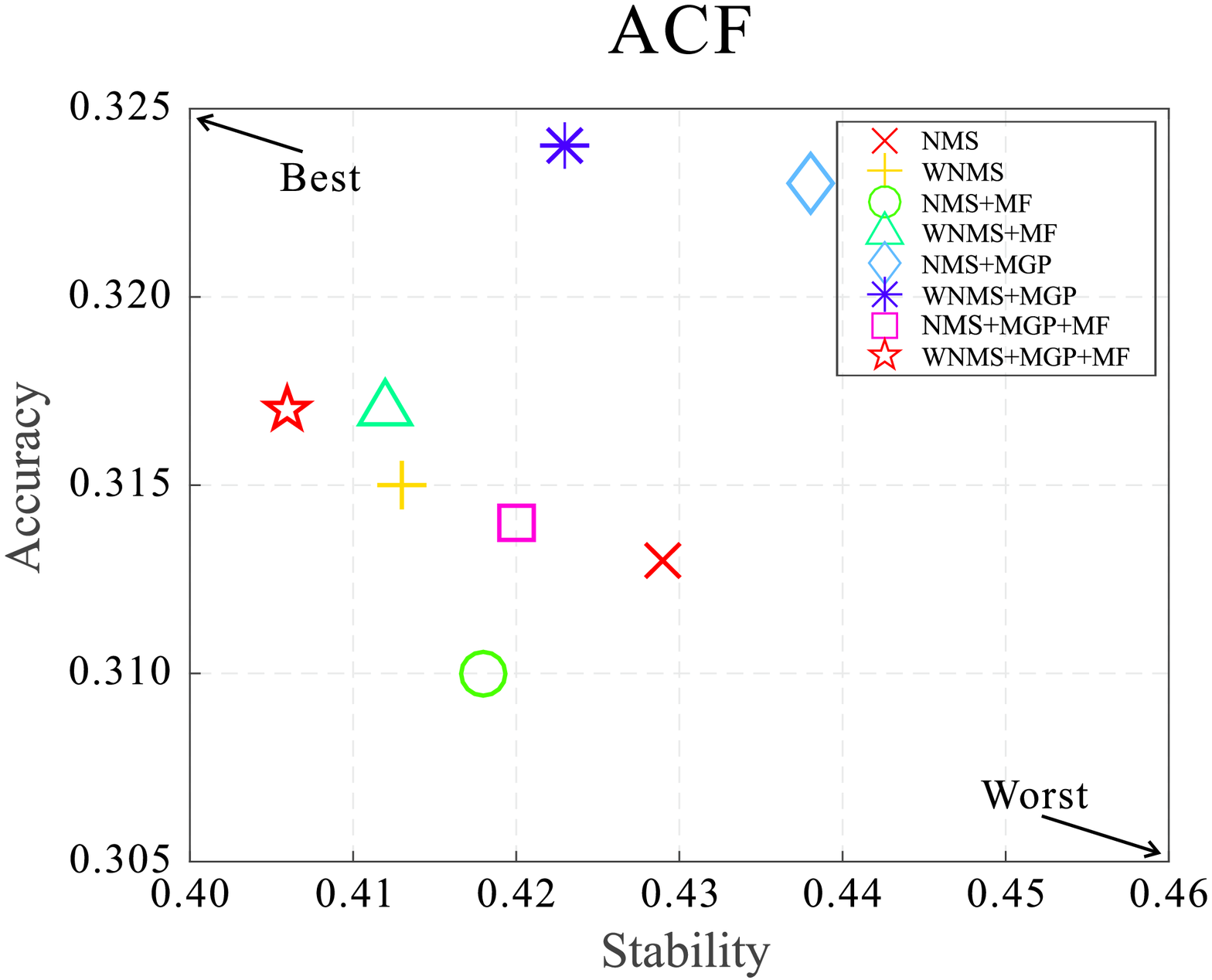}}
    \subfigure[Faster RCNN on KITTI]{\includegraphics[width=0.48\linewidth]{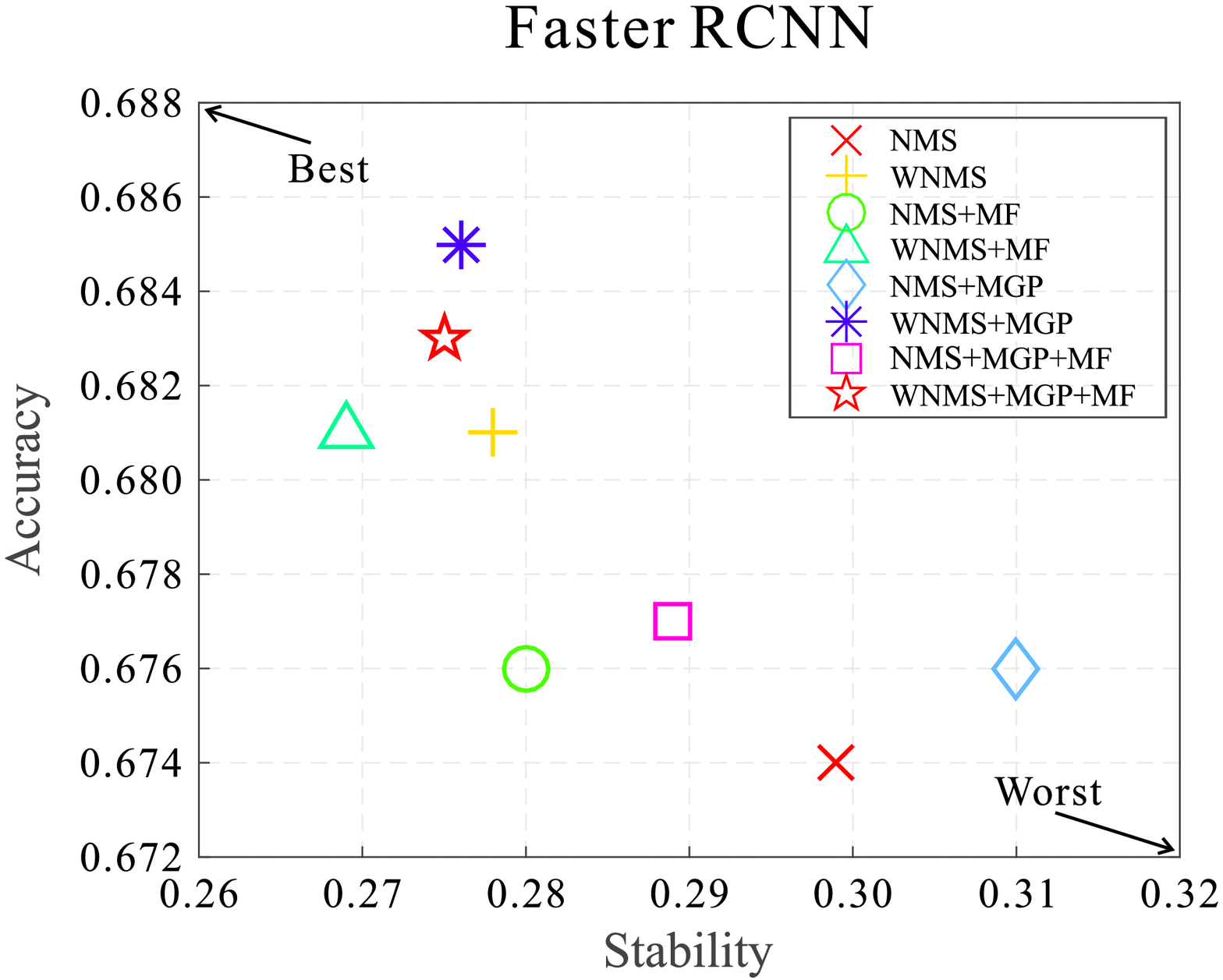}}
    \caption{An accuracy-stability visualization for all methods on MOT and KITTI detected by ACF and Faster RCNN}
    \label{fig:corr_sc}
\end{figure}

In order to interpret the trade-off between accuracy and stability, we draw the scatter plot of accuracy and stability of various methods in Fig.~\ref{fig:corr_sc}. The values are mAUC of accuracy curve and stability curve, respectively. The trends on these two datasets of two detectors are similar. Interestingly, we find that there is no single best method that outperforms others in both accuracy and stability. Specifically, weighted NMS boosts both these two metrics, while MF mostly improves the stability and MGP improves the accuracy. Although MF and MPG both utilize motion information to guide detection, the impact on performance differs and complements. In practice, how to compromise between these two aspects relies on the application at hand.

\section{Discussion and Future Works}
One limitation of our work is that we need ground-truth associations to evaluate the stability. 
However, it usually takes a long time to annotate each trajectory in the videos. We will try to investigate how to reduce such labor works in our future work.

Currently, most VID methods rely on the explicit associations to refine the final results. This kind of methods actually blurs the boundary of VID and MOT: They can also output the associations between detections easily. However, this is not the only way to consider temporal context. Very recently, two concurrent works~\cite{valipour2016recurrent,fayyaz2016stfcn} tried to use the Conv-LSTM framework~\cite{convlstm} to address the temporal continuity in video segmentation. They could enjoy all benefits of end-to-end learning without explicitly calculating motion or associations between frames. Nevertheless, seeking such a method for VID is still an open problem. Last but not least, though we have proposed a metric to evaluate the stability in VID, we still cannot model it directly into learning. How to integrate the stability error into VID and MOT formulation is also an interesting direction to pursue.


\section{Conclusion}

In this paper, we have investigated one important missing component in current VID and MOT evaluation -- stability. First, we analyzed the sources of the instability, and further decomposed it into three terms: fragment error, center position error, scale and ratio error. For each term, we proposed its corresponding evaluation metric. These proposed metrics are intuitive and easy to measure. Next, we conducted comprehensive experiments to evaluate several existing methods that improve over still image detectors based on the new metrics. Through the empirical analyses, we have justified that accuracy and stability are complementary. Both metrics are necessary to characterize the performance of a detector in VID. Furthermore, we have demonstrated the trade-off between accuracy and stability for existing VID methods. We wish our work could inspire more subsequent works that address the stability issue in VID.


{\small
\bibliographystyle{ieee}
\bibliography{egbib}
}

\end{document}